\documentclass{SCIS2026}
\newcommand{\methodname}{MNAFT}

\newcommand{\llavanext}{LLaVA-NeXT}

\newcommand{\clip}{CLIP}
\newcommand{\qwen}{Qwen2.5-VL}
\newcommand{\llama}{LLaMA3}
\newcommand{\internvl}{InternVL2}
\newcommand{\mllama}{LLaMA3.2}

\newcommand{\datasetmit}{MIT-10M}
\newcommand{\datasetiitm}{IIMT}
\newcommand{\datasetecoit}{ECOIT}
\newcommand{\datasetopus}{OPUS-MIT-5M}

\usepackage{caption}
\usepackage{latexsym}
\usepackage{algorithm}
\usepackage{algorithmic}
\usepackage{graphicx}
\usepackage{booktabs}
\usepackage{algorithm}
\usepackage{algorithmic}
\usepackage{amsfonts,amssymb}
\usepackage{amsmath,bm}
\usepackage{subcaption}
\usepackage{multirow} 
\usepackage{multicol} 
\usepackage{xcolor}
\usepackage{colortbl}
\definecolor{Gray}{gray}{0.93}
\usepackage{mathrsfs}
\usepackage{tabularx}
\usepackage{makecell}
\usepackage{amsfonts}
\usepackage{bbm}
\usepackage{arydshln}
\usepackage{booktabs}
\usepackage{bbding}
\usepackage{pifont}
\begin{document}
%\oa
%%%%%%%%%%%%%%%%%%%%%%%%%%%%%%%%%%%%%%%%%%%%%%%%%%%%%%%
%%% Authors do not modify the information below
%%% 作者不需要修改此处信息
\ArticleType{RESEARCH PAPER}
%\SpecialTopic{}
\Year{2025}
\Month{January}
\Vol{68}
\No{1}
\DOI{}
\ArtNo{}
\ReceiveDate{}
\ReviseDate{}
\AcceptDate{}
\OnlineDate{}
\AuthorMark{}
\AuthorCitation{}
%%%%%%%%%%%%%%%%%%%%%%%%%%%%%%%%%%%%%%%%%%%%%%%%%%%%%%%

%%% title: 标题
%%%   \title{title}{title for citation}
% \title{Title}{Title for citation}
\title{\methodname: Modality Neuron-Aware Fine-tuning of  Multimodal Large Language Models for Image Translation}{\methodname: Modality Neuron-Aware Fine-tuning of  Multimodal Large Language Models for Image Translation}

%%% Corresponding author: 通信作者
%%%   \author[number]{Full name}{{email@xxx.com}}
%%% General author: 一般作者
%%%   \author[number]{Full name}{}
%%% Equal Contribution: 同等贡献作者
%%%   \author[number\dag]{Full name}{}
% \author[1]{Aaa AUTHOR}{}
% \author[1,2]{Bbb AUTHOR}{{bauthor@xxx.com}}
% \author[2]{Ccc AUTHOR}{}
% \author[3]{Ddd AUTHOR}{}
\author[2,5]{Bo LI}{}
\author[3]{Ningyuan DENG}{}
\author[1]{Tianyu DONG}{}
\author[4]{Shaobo WANG}{}
\author[1]{Shaolin ZHU}{{zhushaolin@tju.edu.cn}}
\author[2]{Lijie WEN}{{wenlj@tsinghua.edu.cn}}

%%% Authors' contribution. 同等贡献声明
%\contributions{These authors contributed equally to this work.}

%%% Address. 地址
%%%   \address[number]{Affiliation, City Postcode, Country}
% \address[1]{Affiliation, City 000000, Country}
% \address[2]{Affiliation, City 000000, Country}
% \address[3]{Affiliation, City 000000, Country}
\address[1]{School of Computer Science and Technology, Tianjin University, Tianjin, China}
\address[2]{School of Software, Tsinghua University, Beijing, China}
\address[3]{School of Information Resource Management, Renmin University of China,Beijing, China}
\address[4]{School of Artificial Intelligence, Shanghai Jiao Tong University, Shanghai, China}
\address[5]{Baidu Inc., Beijing, China}

%%% Abstract. 摘要
\abstract{
Multimodal Large Language Models (MLLMs) have shown impressive capabilities, yet they often struggle to effectively capture the fine-grained textual information within images crucial for accurate image translation. 
This often leads to a modality gap between visual text inputs and textual inputs/outputs for image translation.
Existing methods, primarily relying on instruction fine-tuning, risk parameter redundancy of pre-trained knowledge, hindering generalization performance. 
To address this, we introduce Modality Neuron-Aware Fine-tuning (\methodname), a novel approach that takes advantage of the specialized roles of individual neurons within MLLMs for enhanced image translation. 
\methodname~identifies language-agnostic and language-specific neurons in both vision and language modules through an instruction-driven activation analysis, evaluating their importance in various translation tasks. 
We then perform selective fine-tuning, updating only the parameters of language-specific and language-agnostic neurons within the selected layers relevant to the target task, while preserving the knowledge encoded in other neurons and layers. 
Our extensive experiments on multiple benchmarks demonstrate that \methodname~significantly outperforms state-of-the-art image translation methods, including cascaded models, standard full fine-tuning, and parameter-efficient tuning techniques. 
Furthermore, we provide comprehensive analysis, including visualizations of neuron activations and clustering patterns, to offer insights into the roles of different neuron groups in mediating cross-modal understanding and facilitating accurate language-specific translation.
}

%%% Keywords. 关键词
% \keywords{keyword1, keyword2, keyword3, keyword4, keyword5}
\keywords{Vision-Language Models, Multilingual Image Translation, Large Language Models }

\maketitle

%%%%%%%%%%%%%%%%%%%%%%%%%%%%%%%%%%%%%%%%%%%%%%%%%%%%%%%
%%% The main text. 正文部分
%%%%%%%%%%%%%%%%%%%%%%%%%%%%%%%%%%%%%%%%%%%%%%%%%%%%%%%

\section{Introduction}
\label{sec:intro}

Multimodal Large Language Models (MLLMs) have recently revolutionized the landscape of artificial intelligence, demonstrating remarkable capabilities in tasks that require a unified understanding of both visual and textual information, such as visual question answering, image captioning, and visual reasoning \cite{dai2023instructblip,li2023blip,wu2023multimodal,zhang2024mm,geminiteam2024gemini15unlockingmultimodal,hong2024cogvlm2,xue2024xgen}.
These models, typically built by integrating a pre-trained visual encoder with a powerful Large Language Model (LLM) via a connector module, represent a significant advance towards building general-purpose AI systems. 
However, a crucial frontier for MLLMs lies in their application to complex, real-world tasks that demand fine-grained understanding and manipulation of multimodal information. 
% Image translation (IT), the task of automatically converting an image containing text in a source language into a visually similar image to the text rendered in a target language, is precisely such a challenge \cite{zhu-etal-2023-peit}.
Image Translation (IT), which aims to translate text embedded in images from the source language to the target language, is precisely such a challenge \cite{zhu-etal-2023-peit,lan2023exploring,watanabe1998translation,yang2002automatic,afli-way-2016-integrating}.

The ability of IT has profound implications for the global accessibility of visual content, cross-cultural communication, and international commerce, making it a critical area of research \cite{lan-etal-2024-translatotron}. 
Traditional approaches to IT, often employed in systems like Google Translate's Instant Camera and Google Lens, rely on a cascaded pipeline of optical character recognition (OCR) followed by machine translation (MT). 
This approach suffers from error propagation (OCR inaccuracies that impact MT), computational inefficiency (due to sequential processing) and a lack of holistic contextual understanding, since the OCR and MT components operate independently \cite{mansimov2020towards, lan-etal-2024-translatotron,li-etal-2025-mit}.

End-to-end (E2E) image translation models offer a more direct and potentially more accurate solution \cite{jain2021image,Ma2022ImprovingET,ma2023multi,zhu-etal-2023-peit}, but adapting MLLMs to this task effectively requires addressing a fundamental modality gap. 
Visual encoders within MLLMs are typically pre-trained on vast image-text datasets using contrastive learning \cite{bai2023qwen,chen2024internvl,hong2024cogvlm2,liu2024llavanext,yao2024minicpm,lu2024deepseek,li2023blip}. 
While this pre-training is effective for general visual-text understanding, it may not be optimal for capturing the subtle, nuanced characteristics of multilingual text embedded within images, which are critical for high-fidelity image translation. 
This discrepancy between the visual-text input representation and the desired textual output leads to suboptimal translation accuracy and fluency. 
Models such as \internvl~\cite{chen2024internvl}, \llavanext~\cite{liu2024llavanext}, \mllama~\cite{llama3modelcard},  and \qwen~\cite{bai2025qwen2} despite their impressive general capabilities are demonstrably affected by this limitation.

The predominant paradigm for adapting MLLMs to specific tasks is instruction fine-tuning, where the pre-trained model is further trained on task-specific data, often with instructions guiding the model's output. 
Fine-tuning MLLMs for IT \cite{li-etal-2025-mit} is intuitive, but the prevailing practice, exemplified by models such as \llavanext~\cite{liu2024llavanext}, involves uniformly updating all model parameters. 
This approach neglects the functional specialization of individual neurons within MLLMs. 
Neurons are not homogeneous; they develop distinct roles in processing different modalities and languages. 
Uniform updates risk disrupting pre-trained knowledge and are suboptimal for specialized tasks like image translation. 
Although previous work explores network analysis \cite{tang2024language}, it lacks a practical and targeted method of fine-tuning based on neuron specialization to improve multimodal task performance. 
A more refined approach, explicitly leveraging neuron-specific roles, is essential to unlock the full potential of MLLMs for image translation.

%%%%%%%%%%%%%%%%%%%%%%%%%%%%%%%%%%%%%%%%
% \begin{figure}[t]
%   \centering

%    \includegraphics[width=\textwidth]{img/Illustration.pdf}
%   \caption{The Illustration  of our proposed \methodname~Framework. (a) Demonstration of \methodname~Framwork with \qwen-3B. (b) Specific and general neuron analysis of the visual and language layers to identify neurons that require fine-tuning. (c) Selective Fine-tuning Strategy, updating only the parameters of language-specific and language-agnostic neurons within the selected layers relevant to the target task, while preserving the knowledge encoded in other neurons and layers.}
%   \label{fig: illustration}
% \end{figure}
%%%%%%%%%%%%%%%%%%%%%%%%%%%%%%%%%%%%%%%%

To address this issue, we introduce Modality Neuron-Aware Fine-tuning (\methodname), a novel and principled method specifically designed to optimize MLLM for image translation by leveraging the functional specialization of individual neurons. 
% As shown in Figure \ref{fig: illustration},
\methodname~is based on the key insight that different neurons within the vast network of an MLLM develop different roles: some become specialized in processing specific languages, others to handle visual features, and still others to bridge the gap between modalities. 
By accurately identifying and selectively targeting these neurons during fine-tuning, we can achieve significantly improved translation performance while preserving the general capabilities of the MLLM.
\methodname~consists of two core stages: (I) We introduce a robust methodology with instruction-driven, using both activation patterns and gradient information, to identify language-specific and modality-shared neurons within the visual and textual processing pathways of the MLLM. 
This creates a precise ``functional map" of neuron specialization within the network.
(II) Guided by the identification of neurons, we implement a selective fine-tuning strategy. 
During fine-tuning on target language image translation data, only the parameters of language-specific and language-agnostic neurons within the selected layers relevant to the target task are updated. 
The remaining neurons are frozen, preventing disruption of pre-trained knowledge and avoiding parameter redundancy.
This targeted approach maximizes fine-tuning efficiency and minimizes unintended consequences.

We perform a rigorous evaluation of \methodname~across a comprehensive suite of image translation benchmarks, encompassing diverse datasets and translation tasks. 
Our experimental results demonstrate that \methodname~substantially outperforms state-of-the-art baselines, establishing a new performance benchmark for the field. 
Moreover, we provide an in-depth analysis, including novel visualizations, that elucidates the functional roles of different neuron groups in mediating cross-modal understanding and achieving accurate, language-specific translation.
This analysis provides unprecedented insights into the inner workings of MLLMs, contributing significantly to our fundamental understanding of these powerful models.

Our contributions can be summarized as follows.
\begin{itemize}
    \item We introduce \methodname, the first modality neuron-aware fine-tuning method specifically designed to optimize MLLM for image translation, achieving superior performance and efficiency.
    \item We present a detailed analysis with novel visualizations that reveal the functional specialization of neurons within MLLMs during cross-modal and language-specific processing, significantly advancing our understanding of these complex models.
    \item We demonstrate, through comprehensive experiments, that \methodname~achieves state-of-the-art results in image translation, setting a new benchmark for the field.
    % \item \textbf{Broad Potential}: The MNAFT method is not restricted to the image translation and has general applicability to other multimodal tasks.

\end{itemize}

\section{Related work}
\label{sec:related_work}
\subsection{Image translation}

Image Translation (IT) aims to translate texts embedded in images from the source language to the target language \cite{lan2023exploring}.
Its wide range of applications makes it a valuable field of research \cite{yang2002automatic,afli-way-2016-integrating}.
Current image translation systems can be divided into two paradigms: cascaded and end-to-end approaches.
Cascaded approaches employ an OCR to extract text from the input image, followed by a separate Neural Machine Translation (NMT) for translation \cite{goodfellow2013multi,zhang2016variational,gu2017non}.
However, this approach suffers from error propagation, massive parameters, and complexity of deployment \cite{mansimov2020towards, lan-etal-2024-translatotron}.
Eventually, end-to-end IT that integrates OCR and MT modules into a single model has attracted much attention \cite{jain2021image}. 
\cite{Ma2022ImprovingET} applies multi-task learning
to this task, where NMT and OCR are jointly trained.
Furthermore, \cite{ma2023multi} apply knowledge distillation to effectively distill the knowledge of OCR and NMT into end-to-end IT.
\cite{zhu-etal-2023-peit} explore an end-to-end IT with an aligner and a regularizer to reduce the modality gap, and \cite{lan2023exploring} introduce an IT model with multimodal codebooks to reduce the impact of OCR errors. 
\cite{lan-etal-2024-translatotron} use a target text decoder and an image tokenizer to alleviate the language alignment burden and improve performance by transforming long pixel sequences into shorter visual token sequences.
Recently,  MLLMs \cite{gpt-4o,liu2024llavanext,lu2024deepseek,llama3modelcard,geminiteam2024gemini15unlockingmultimodal,li2023blip,hong2024cogvlm2,xue2024xgen} have demonstrated impressive performance in various tasks such as visual question answering, visual understanding, and reasoning.
These solutions normally follow to utilize the visual encoder to encode visual features and utilize the connector module to project visual tokens into the word embedding space of the LLM.
However, the visual encoder (e.g. \clip), which is primarily pre-trained on image-text pairs with contrastive learning. 
Therefore, it is an intuitive solution to fine-tune MLLM to enhance performance on the IT task \cite{li-etal-2025-mit}.

\subsection{Multimodal large language model instruction tuning}

Instruction tuning has significantly improved the generalization capability of Multimodal Large Language Models (MLLMs) on various tasks \cite{dai2023instructblip,li-etal-2025-mit,wei2021finetuned,ouyang2022training}. 
However, standard full fine-tuning updates a large number of weights in all intermediate layers and the pre-trained LLM, leading to parameter redundancy and high computational costs \cite{xu2024pllava,zhang2024ai}.
Therefore, Parameter-Efficient Fine-Tuning (PEFT) methods have been proposed \cite{hu2022lora,yan2023prompt,jie2023revisiting,he2023parameter}. Among them, Low-Rank Adaptation (LoRA) \cite{hu2022lora} and its variants such as DORA \cite{pmlr-v235-liu24bn} have become widely accepted as PEFT methods, where the fine-tuning of models is performed by updating a small number of injected adaptation parameters.
However, in multimodal instruction tuning, traditional PEFT methods often suffer from parameter redundancy, as they rely on fitting a limited number of common parameters to perform different tasks, severely affecting the transferability between previously learned datasets \cite{french1999catastrophic,kirkpatrick2017overcoming,luo2023empirical,zhai2023investigating}. 
Moreover, most existing PEFT methods focus on single modalities and neglect the crucial role of multimodal features in fine-tuning \cite{wang2022dualprompt,ju2022prompting,han20232vpt}. For example, some methods typically freeze the visual encoder and only fine-tune the connector layers and the LLM component, limiting the model's ability to fully utilize multimodal information.
To address these limitations, MixLoRA\cite{shen-etal-2024-multimodal} builds on LoRA by dynamically constructing low-rank adaptation matrices tailored to the unique requirements of each input, with the goal of mitigating task interference. 
M$^2$PT \cite{wang-etal-2024-m2pt} facilitates cross-modal feature extraction and matching of cross-modal features by injecting visual and textual prompts into the visual and textual processors, respectively. 
% SPIDER proposes a refinement method based on importance discrepancy evaluation. Here, parameters are selectively updated based on the measured importance of pre-trained weights and fine-tuning of gradients, which avoids parameter redundancy .
In contrast to the aforementioned methods, our proposed Modality Neuron-Aware Fine-tuning (\methodname) method explicitly considers the effects of both cross-modal interactions and language-specific neuron behavior on image translation. \methodname~first identifies language-specific and general neurons within the model by analyzing their activation patterns and gradient information during cross-modal interactions. During fine-tuning for a target language, \methodname~then selectively updates only the relevant cross-modal and language-specific neurons while freezing others, thus preserving knowledge about other languages and modalities. This targeted strategy minimizes interference with the translation capabilities of other languages and improves generalization performance.

\section{Preliminaries}
\label{sec:Preliminaries}

\subsection{Task definition}
Formally, we define the task of image translation (IT) using a dataset $\mathcal{D} = \{ (v_i, s_i, t_i) \}_{i=1}^N$, where
$v_i$ denotes the input image with the text of the source language.
% $s_i$ denotes the text string of the source language extracted from $v_i$. 
% % Although we used this text to calculate the importance of neurons, \methodname~works directly with the image during inference.
% $t_i$ denotes the corresponding target language translation of $s_i$.
$s_i$ denotes the text string of the source language extracted from $v_i$, which serves as auxiliary information for neuron analysis. 
$t_i$ denotes the corresponding target language translation of $s_i$. 
The goal of IT is to generate the optimal target translation $t_i$ directly from the input image $v_i$.
% The goal of IT is to generate the optimal target translation $t_i$ for the input image $v_i$. 
During training, we can express the loss function as follows:
\begin{equation}
\mathcal{L}_{IT} = - \frac{1}{N} \sum_{i=1}^{N} \log p(t_i | v_i; \theta) ,
\end{equation}
where $\theta$ denotes the parameters of the MLLM.

\subsection{Taylor expansion}

The Taylor expansion provides a polynomial approximation to a function \cite{molchanov2016pruning,Xie2021ImportancebasedNA,zhu-etal-2023-peit}. 
We use it to analyze how the impact of removing a neuron affects the loss of the model. 
For a given neuron $i$ with activation $h_i$, we examine the effect of setting $h_i = 0$ on the loss function $\mathcal{L}(H, h_i)$, where $H$ denotes the activations of all other neurons.
The Taylor expansion of $\mathcal{L}(H, h_i)$ by one point $a$ is as follows:
\begin{equation}
\mathcal{L}(H, h_i) = \sum_{n=0}^{\infty} \frac{\mathcal{L}^{(n)}(H, a)}{n!} (h_i - a)^n ,
\end{equation}
where $\mathcal{L}^{(n)}(H, a)$ is the $n$-th derivative of $\mathcal{L}$ with respect to $h_i$, evaluated with respect to $a$. 
For practical purposes, we use the first-order approximation:
\begin{equation}
\mathcal{L}(H, h_i) \approx \mathcal{L}(H, a) + \frac{\partial \mathcal{L}(H, a)}{\partial h_i} (h_i - a) .
\end{equation}
To analyze the impact of removing the neuron $i$, we set $a = 0$:
\begin{equation}
\mathcal{L}(H, h_i) \approx \mathcal{L}(H, 0) + \frac{\partial \mathcal{L}(H, 0)}{\partial h_i} h_i .
\end{equation}
The change in loss $\Delta \mathcal{L}(h_i)$ due to the distance of the neuron is then approximated as follows.
\begin{equation}
\Delta \mathcal{L}(h_i) = \mathcal{L}(H, 0) - \mathcal{L}(H, h_i) \approx -\frac{\partial \mathcal{L}(H, 0)}{\partial h_i} h_i .
\end{equation}
The magnitude of this change, $|\Delta \mathcal{L}(h_i)|$, reflects the importance of the neuron. We define the importance value $\Theta_{TE}(i)$ as:
\begin{equation}
\Theta_{TE}(i) = \left| \frac{\partial \mathcal{L}(H, 0)}{\partial h_i} h_i \right| ,
\end{equation}
where:
$|h_i|$ denotes the magnitude of activation of the neuron, reflecting its contribution to the output of the network.
$\left| \frac{\partial \mathcal{L}(H, 0)}{\partial h_i} \right|$ denotes the sensitivity of the loss to changes in neuronal activation, evaluated at $h_i = 0$.
A higher $\Theta_{TE}(i)$ denotes a greater importance. 
This first-order Taylor expansion provides an efficient and effective way to estimate the importance of neurons for our selective fine-tuning strategy.

\section{Method}
%%%%%%%%%%%%%%%%%%%%%%%%%%%%%%%%%%%%%%%%
\begin{figure}[t]
  \centering

   \includegraphics[width=0.85\textwidth]{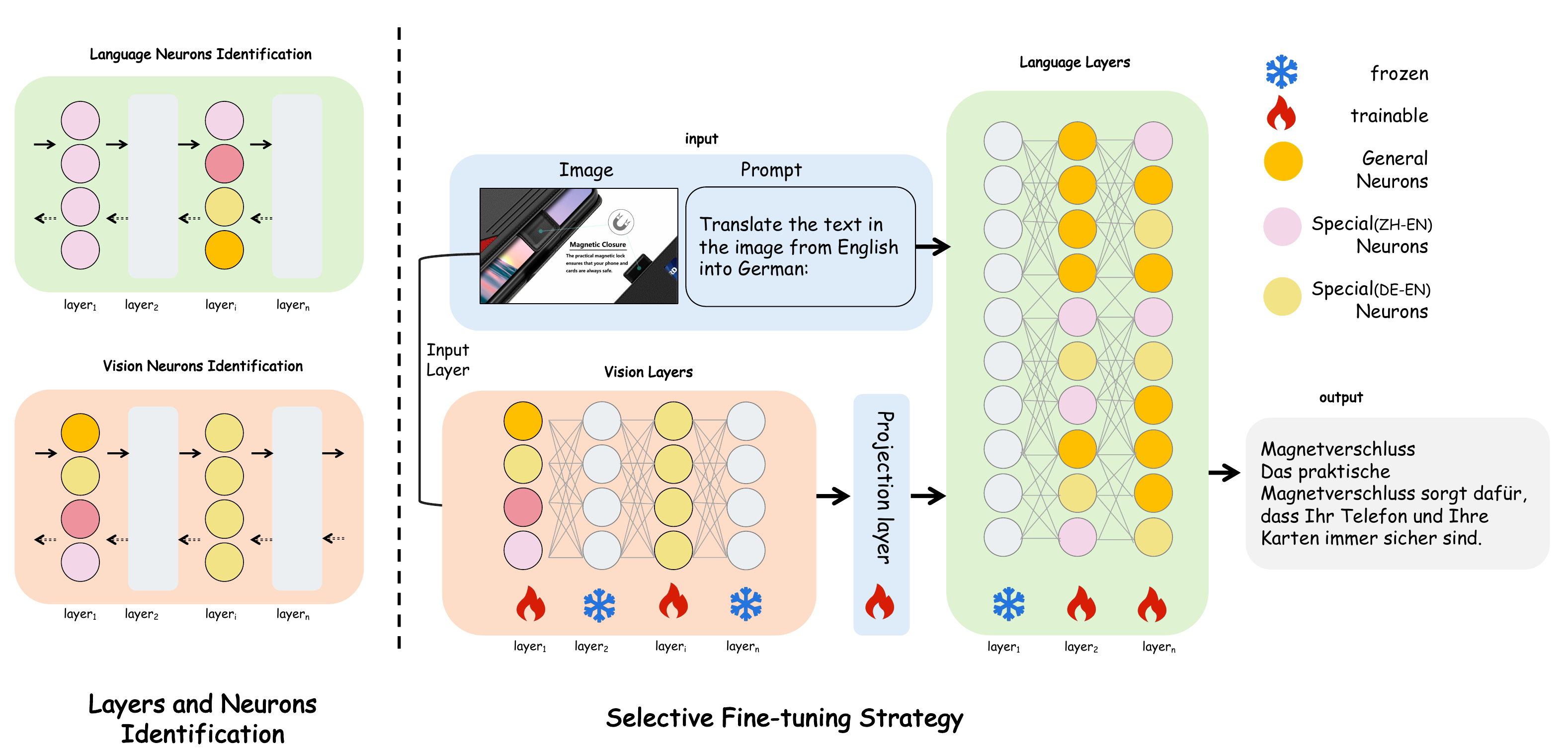}
  \caption{The overview of \methodname~Framework. We begin by evaluating the importance of neurons within both the vision layers and language layers across different languages using a novel instruction-driven approach. This yields neuron importance scores that reflect their contribution to the image translation task. We then select the most influential neurons within key layers of both modules and categorize them into language-agnostic (general) and language-specific groups. Finally, we perform selective fine-tuning by updating only the specific neurons relevant to the target task and the general neurons in the same layer to obtain the encoded general knowledge.}
  \label{fig: fig1_main_method}
\end{figure}
%%%%%%%%%%%%%%%%%%%%%%%%%%%%%%%%%%%%%%%%

Figure~\ref{fig: fig1_main_method} illustrates the key components of our \methodname~methodology.
We begin by evaluating the importance of neurons within both the vision layers and language layers across different languages using a novel instruction-driven approach. 
This yields neuron importance scores that reflect their contribution to the image translation task. 
We then select the most influential neurons within key layers of both modules and categorize them into language-agnostic (general) and language-specific groups. 
Finally, we perform selective fine-tuning by updating only the specific neurons relevant to the target task and the general neurons in the same layer to obtain the encoded general knowledge.
This targeted approach mitigates parameter redundancy  and parameter interference.

% \subsection{Neurons Awareness Evaluation}
\subsection{Insight}

Different neurons within MLLMs exhibit varying degrees of specialization for different languages and visual modalities, posing challenges for standard fine-tuning. 
Indiscriminate fine-tuning can lead to:
(I) erosion of general knowledge essential for robust multimodal understanding, and (II) interference between language-specific features. 
\methodname~addresses this by employing an instruction-driven approach to identify and prioritize relevant neurons for each language and task.
Instead of random perturbations, we use targeted instructions to test the activation of neurons, such as the instruction ``What is the text in the picture?'', which activates neurons responsible for recognizing and processing text within an image, and the instruction ``Translate the text from [Source Language] into [Target Language]. [Source Language]: [Source Text] [Target Language]:'' activates neurons specialized in translating between specific language pairs.

We propose a method of selective neuron adaptation that uses Taylor expansion (TE)-based importance scores to identify and prioritize the most relevant neurons for each language. 
Using Taylor expansion, we can effectively evaluate the importance of each neuron in different image translation tasks. In this context, we define the awareness score, denoted as $\Phi(i)$, for each neuron with respect to a particular task. 
With this innovative approach, we can not only evaluate the importance of individual neurons but also distinguish between the neurons that are crucial for the processing of the entire task and the neurons that exert a particularly strong influence within a specific task.
\begin{equation}
\Phi(i) = |\Delta L(h_i)|, \quad i \in L_j,
\end{equation}
where $ L_j$ denotes the layer previously selected. $h_i$ denotes the output generated by the neuron $i$, while $|\Delta L(h_i)|$ refers to the corresponding loss value induced by perturbing specific neurons.
By computing $\Phi(i)$across multiple tasks and language pairs, we construct a matrix of importance of neurons. 
This matrix reveals which neurons are consistently important across tasks (language-agnostic) and which specialize in particular languages or tasks (language-specific).

To establish a clear and explicit correlation between the neuronal activation value and its effect on the loss function, we utilize the Taylor expansion, which can be expressed as follows:
\begin{equation}
|\Delta L(h_i)| = \left| \frac{\partial L}{\partial h_i} h_i \right|.
\end{equation}
This refined approach allows us to systematically evaluate the importance of neurons and prioritize those neurons that significantly contribute to translation accuracy. 
Unlike conventional magnitude-based pruning methods, which often overlook the intricate relationships between neuron activations and linguistic relevance, our gradient-aware significance metric demonstrates superior performance. 
By focusing on the nuanced contributions of individual neurons, we can enhance the overall effectiveness of image translation systems, ensuring that they remain robust and accurate in diverse linguistic contexts. 

% \subsection{Relevant Module Selection}
\subsection{Neurons identification}

Following the scoring process, we perform a thorough analysis of the layers by aggregating the neuron scores. 
This aggregation allows us to identify neurons that exhibit significant relevance to the task at hand. 
Understanding these significant neurons is crucial for revealing the modular characteristics inherent in the neural network architecture, as it provides insights into how different layers contribute to overall performance. 
To quantify the relevance of each neuron within a specific layer, we define the following equation.
\begin{equation}
\Theta_m^{\text{TE}}(i_l) = \frac{1}{T_m} \sum_t \left| \frac{\delta L(H, h_{li})}{\delta h_{li}} h_{l}^{i} \right|.
\end{equation}
In this equation, $m$ denotes the total number of neurons within each layer and $\Theta_m^{\text{TE}}(i_l)$ serves as the relevance score for that specific layer. 
This score reflects the impact of each neuron on the overall loss function, providing a clear metric for assessing their importance.
To enhance the model's ability to focus on specific languages, particularly those that are less represented or possess unique linguistic features, we implemented a normalization process across the layers. 
This normalization is crucial for several reasons. It helps mitigate the impact of inherent biases that may arise from the training data, allowing the model to achieve a more balanced understanding of different languages.
% Our approach not only improves the model's efficacy, but also contributes to a more inclusive and equitable representation.

Next, we rank the layers according to their calculated relevance scores, organizing them from highest to lowest importance. 
This ranking process is essential for selecting the most significant modules within both the vision layers and the language layers, allowing us to focus our optimization efforts on where they will be most effective. 
The selection of the top layers can be expressed mathematically as follows.
\begin{equation}
L_{vision}=arg   \mathop{\max}_{top_{l}} \left\{D_{1}, D_{2}, D_{k_{vision}}    \right\} ,
\end{equation}
\begin{equation}
L_{LLM}=arg   \mathop{\max}_{top_{l}} \left\{  D_{1},D_{2},D_{k_{LLM}}    \right\} ,
\end{equation}
where $L_{vision}$ and $L_{LLM}$ represent the selected layers of the vision encoder and the LLM, respectively, while $ D_{1}, D_{2}, D_{k_{vision}}$ and $D_{1}, D_{2}, D_{k_{LLM}}$ denote the relevance scores for the respective layers. 
We selected $L_{vision}$ and $L_{LLM}$ as the most critical layers, with $l_{vision}$ and $l_{LLM}$ serving as hyperparameters. 
This structured approach enables us to effectively identify and select the most impactful modules for further optimization.

After identifying the important layers, we proceed to rank the neurons within each layer based on their scores derived from the Taylor expansion. 
This ranking process is crucial, as it allows us to pinpoint the most influential neurons that contribute significantly to the overall task performance. 
To analyze a single layer effectively, we sort the neurons according to their variance, which provides insight into their individual contributions:
\begin{equation}
\lambda(i) = \text{sort}(\sigma(X_i))\left\lfloor \epsilon \times p \right\rfloor, \quad i \in L,
\end{equation}
where $p$ denotes the total number of neurons in the $i$ layer, while $\epsilon$ is a predefined ratio that helps establish a threshold for classification. 
Neurons exhibiting a variance in the linguistic awareness score below the estimated threshold $\lambda(i)$ are categorized as neurons of general language, indicating their greater applicability across multiple languages. 
In contrast, those that exceed this threshold are classified as specific language neurons, highlighting their specialized role in understanding particular linguistic features.

To further refine our approach, we aggregated the neurons into distinct collections customized to each pair of languages. 
This targeted aggregation allows us to optimize the model's performance by focusing on the unique characteristics and nuances of the languages involved. 
By leveraging the insights gained from this ranking and classification process, we can enhance the model's ability to process and generate language more effectively, ultimately leading to improved outcomes in various linguistic tasks.

\subsection{Selective fine-tuning strategy}

\methodname~employs a highly selective fine-tuning strategy targeting specific neurons within the identified layers of the vision and language modules. 
This approach preserves the general knowledge captured by the pre-trained model while enabling efficient adaptation to the target task. 
The selection process, guided by neuron importance scores and variance analysis, results in two key sets of neurons for each relevant layer $l$: language-agnostic neurons and language-specific neurons.

During fine-tuning, we freeze all parameters except the weights and biases of the selected language-specific neurons and the language-agnostic neurons within the identified layers.  
This crucial detail ensures that the general knowledge encoded in language-agnostic neurons is preserved and contributes to overall performance. 
Let $W_l$ and $b_l$ represent the weight matrix and the bias vector of layer $l$, respectively. For each layer $l$ selected for fine-tuning, we apply a mask $M_l$ to the gradients during backpropagation. This mask is defined as follows:
\begin{equation}
M_l[i] = \begin{cases} 1, & \text{if } i \in A_l \cup S_{l,t} \\ 0, & \text{otherwise} \end{cases},
\end{equation}
where $i$ indexes the neurons in the layer $l$. This mask effectively zeros out the gradients for all neurons except those in $A_l$ and $S_{l,t}$. The update rule for the parameters of layer $l$ becomes:
\begin{equation}
W_l^{t+1} = W_l^t - \alpha \cdot (\nabla_{W_l} L_t \odot M_l),
\end{equation}
\begin{equation}
b_l^{t+1} = b_l^t - \alpha \cdot (\nabla_{b_l} L_t \odot M_l),
\end{equation}
where $W_l^{t+1}$ and $b_l^{t+1}$ are the updated weight matrix and bias vector for layer $l$ at time step $t+1$, $W_l^t$ and $b_l^t$ are the weight matrix and bias vector at time step $t$.
$\alpha$ is the learning rate, $L_t$ is the task-specific loss at time step $t$, $\nabla_{W_l} L_t$ and $\nabla_{b_l} L_t$ are the gradients of the loss with respect to the weights and biases of layer $l$, respectively, $\odot$ denotes element-wise multiplication. 
This selective application of gradients ensures that only the chosen language-specific and language-agnostic neurons are updated during fine-tuning, preserving general knowledge while adapting to the specific target task.

\section{Experiments}
We conducted extensive experiments on six tasks in four publicly available image translation datasets to demonstrate the effectiveness of our proposed \methodname~compared to existing end-to-end and cascaded baseline approaches. 
We also performed ablation studies to analyze the contribution of each component within \methodname.

\subsection{Setup}
% \noindent \textbf{Datasets.} 
\subsubsection{Datasets}
We conducted comprehensive experiments on six tasks in four publicly available image translation datasets.
There are two synthetic and two real datasets.

\begin{itemize}
    \item \textbf{\datasetecoit} \cite{zhu-etal-2023-peit} is a large-scale image translation dataset in the e-commerce domain, containing product images automatically crawled from a Chinese e-commerce website\footnote{\url{https://www.taobao.com/}} paired with post-edited target translations (480K sentences with 3.64M source tokens).
    
    \item \textbf{\datasetiitm} \cite{lan-etal-2024-translatotron} utilizes the IWSLT14 \cite{cettolo2014report} German to English dataset and the IWSLT17 \cite{cettolo2017overview} French to English dataset to synthesize paired images. In addition, the background color of the image is selected randomly and the resolution of the images is 512×512. 
    It comprises 452,230 instances.
    
    \item \textbf{\datasetmit} \cite{li-etal-2025-mit} is a large-scale parallel corpus of multilingual image translation with over 10M image-text pairs derived from real-world data, which has undergone extensive data cleaning and multilingual translation validation. It contains 840K images and 14-languages image-text pairs.
    
    \item \textbf{\datasetopus} \footnote{\url{https://huggingface.co/datasets/liboaccn/OPUS-MIT-5M}} is constructed by randomly sampling 5M sentence pairs from the OPUS corpus. This image translation dataset contains 5 million sentence pairs and 20 language pairs.
\end{itemize}

For each task, we constructed a training set by 100K items from the source dataset. 
The evaluation was performed on a test set of 100 items per task. 
Notably, the neuron identification stage of \methodname~was also conducted using the test set.
The neuron identification stage does not involve any parameter updates; it only computes importance scores based on activation magnitudes and gradient information to determine the structural selection of neurons and layers. 
This process is analogous to calibration-based pruning, where a small reference set is used to assess neuron saliency without optimizing model weights. Since no learning occurs during this stage, the use of the test set does not introduce evaluation bias or test-set leakage.
%%%%%%%%%%%%%%%%%%%%%%%%%%%%%%%%%%%%%%%%%%
\begin{figure}[t]
  \centering
   \includegraphics[width=0.85\textwidth]{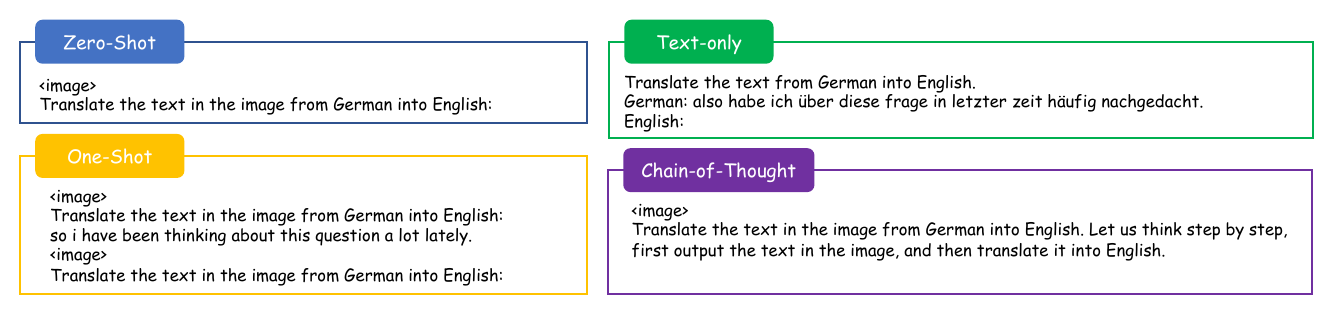}
  \caption{Prompt templates. We used the \qwen-3B model and evaluated its performance using four strategies: Text-Only (text-only translation based on the text extracted from the images), zero-shot, one-shot, and chain-of-thought (CoT) prompting. }
  \label{fig: prompt_template}
\end{figure}
%%%%%%%%%%%%%%%%%%%%%%%%%%%%%%%%%%%%%%%%%%

% \noindent \textbf{Baselines.}  
\subsubsection{Baselines}
In this work, we compared \methodname~with three categories of baseline methods and with six SOTA IT models:

% \begin{itemize}
    \textbf{Cascaded Models}: We used EasyOCR \footnote{https://github.com/JaidedAI/EasyOCR} and PP-OCRv3 \cite{li2022pp} for text extraction from images in the datasets, followed by NLLB-200 \cite{nllb2022} for text translation.
    It is a machine translation model primarily intended for machine translation research, especially for low-resource languages. 
    It allows single-sentence translation between 200 languages.
    
    \textbf{MLLM Baselines Models}: We used the \qwen-3B \cite{bai2025qwen2} model and evaluated its performance using four strategies: Text-Only (text-only translation based on the text extracted from the images), zero-shot, one-shot, and chain-of-thought (CoT)  \cite{NEURIPS2022_9d560961} prompting. 
    The prompt templates are shown in Figure~\ref{fig: prompt_template}.
    
    \textbf{Fine-Tuning Methods}: We compared \methodname~with established fine-tuning techniques: Full Fine-Tuning and four SOTA parameter-efficient fine tuning methods (PEFT):
    \begin{itemize}
    \item Full Fine-Tuning: The default approach, where both the visual and the language layers are updated during fine-tuning.
    \item LoRA \cite{hu2022lora}: Efficiently fine-tunes large pre-trained models by using low-rank decomposition. It adds trainable low-rank matrices to the model's weight matrices, reducing the number of parameters that require fine-tuning. In our experiments, we configured LoRA with $rank=8$ and applied it to all linear layers of the model ($target\_modules=all$).
    \item DoRA \cite{pmlr-v235-liu24bn}: Builds upon LoRA by decomposing pre-trained weights into ``magnitude'' and ``direction'' components for finer-grained control and learning. DoRA optimizes these magnitude and direction components separately.
    \item MixLoRA \cite{shen-etal-2024-multimodal}: A novel approach that integrates multimodal instruction tuning with Conditional Mixture of LoRA. Dynamically constructs low-rank adaptation matrices tailored to each input, instance by selecting their decomposition factors from two collections. 
    \item M$^2$PT \cite{wang-etal-2024-m2pt}: A multimodal prompt-tuning method for efficient instruction tuning of MLLMs. Introduces two sets of soft prompts: visual prompts and textual prompts, which are prefixed to the visual and instruction inputs, respectively.
    \end{itemize}

    \textbf{SOTA IT Models}: As for the end-to-end model, we compare it with mainstream IT models:
    \begin{itemize}
    \item ItNet \cite{jain2021image} is an end-to-end image translation system. It first pre-trains a standard Transformer on a text-only parallel dataset. ResNet is used as an image encoder to encode latent semantic representations of images. 
    \item PEIT \cite{zhu-etal-2023-peit} is an end-to-end image translation framework that bridges the modality gap with pre-trained models. 
    \item Translatotron-V \cite{lan-etal-2024-translatotron} is an end-to-end IT model consisting of four modules. In addition to an image encoder and
an image decoder, it contains a target text decoder and an image tokenizer.
    \item UMTIT \cite{niu-etal-2024-umtit} first encodes the image using a vision transformer and then decodes the translation with a text transformer in an autoregressive manner.
    \item E2ETIT \cite{Ma2022ImprovingET} builds a novel modal adapter that effectively fuses the OCR encoder and the MT decoder. 
    \item DIMTDA \cite{liang-etal-2024-document} is document image machine translation with dynamic multi-pre-trained model assembly.
    \end{itemize}
% \end{itemize}

\subsubsection{Evaluation metrics}
% \noindent \textbf{Evaluation metrics.} 
We evaluated the performance of IT models on several dimensions, including semantic similarity, fluency, and accuracy.
The evaluation metrics are as follows.
\begin{itemize}
    \item BLEU \cite{papineni2002bleu}: It calculates the n-gram overlap between the candidate and reference translations.
    \item METEOR \cite{banerjee-lavie-2005-meteor}: Taking into account synonyms and word order provides a more nuanced assessment of semantic similarity than BLEU.

\end{itemize}

\subsubsection{Implementation Details}
% \noindent \textbf{Implementation Details.}
The operating system that we use is CentOS release 7.5, and the programming language is Python 3.9.12. 
Our experiments were conducted on NVIDIA TESLA A100-80G GPU, the CUDA version is 12.2, and the deep learning framework is torch with version 2.1.0, torchvision with version 0.16.0 and Transformers with 4.45.0.
We used the LlamaFactory \cite{zheng-etal-2024-llamafactory} framework for all fine-tuning experiments. 

% \subsection{Implementation details}
% \subsection{Baselines \& Datasets}
\subsection{Main result}

\begin{table}[t]
    \centering
    \caption{Quantitative comparison with three categories of baseline methods. We conducted a comprehensive evaluation of 11 models or methods (including cascaded, MLLM baseline models and fine-tuning Method) using the test set from 6 language pairs on the 4 datasets. We highlight the best numbers in \textbf{bold}.}
\renewcommand{\arraystretch}{1.1}
\resizebox{0.95\textwidth}{!}{
\begin{tabular}{lcccccccccccc}
\toprule
% & \multicolumn{2}{c}{\datasetecoit} & \multicolumn{4}{c}{\datasetiitm}  & \multicolumn{4}{c}{\datasetmit} & \multicolumn{2}{c}{\datasetopus} \\
\midrule
& \multicolumn{2}{c}{\datasetecoit~(ZH-EN)} & \multicolumn{2}{c}{\datasetiitm~(DE-EN)} & \multicolumn{2}{c}{\datasetiitm~(FR-EN)} & \multicolumn{2}{c}{\datasetmit~(DE-EN)} & \multicolumn{2}{c}{\datasetmit~(EN-DE)} & \multicolumn{2}{c}{\datasetopus~(EN-ZH)} \\
\midrule
Model & METEOR & BLEU & METEOR & BLEU & METEOR & BLEU & METEOR & BLEU & METEOR & BLEU & METEOR & BLEU \\
\midrule
\multicolumn{13}{c}{\textit{Cascaded Models}} \\ \hline
EasyOCR\_NLLB & 13.7 & 8.0 & 42.5 & 22.3 & 50.5 & 31.8 & 5.7 & 1.6 & 21.8 & 11.2 & 43.9 & 33.1 \\
PP-OCRv3\_NLLB & 13.1 & 8.2 & 43.1 & 24.4 & 54.2 & 34.3 & 11.3 & 5.5 & 27.4 & 20.7 & 42.1 & 30.8 \\
\midrule
\multicolumn{13}{c}{\textit{Baseline Models (\qwen-3B)}} \\ 
\midrule
Text-Only & 51.9 & 35.6 & 48.4 & 29.8 & 56.7 & 39.9 & 63.2 & 50.7 & 63.6 & 55.1 & 60.9 & 45.2 \\
Zero-Shot & 44.8 & 28.1 & 46.5 & 27.0 & 49.4 & 31.6 & 28.4 & 19.1 & 31.3 & 24.3 & 63.3 & 46.8 \\
One-Shot & 49.0 & 31.9 & 39.1 & 19.6 & 43.3 & 24.6 & 33.3 & 16.1 & 30.7 & 21.8 & 52.2 & 40.1 \\
Chain-of-Thought & 49.1 & 30.2 & 43.3 & 25.8 & 51.0 & 34.5 & 21.3 & 12.3 & 16.9 & 11.9 & 61.2 & 46.3 \\

\midrule
\multicolumn{13}{c}{\textit{Fine-tuning Method}} \\ 
\midrule
Full Fine-Tuning & 68.5 & 51.4 & 54.6 & 36.5 & 59.9 & 42.3 & 62.6 & 49.6 & 50.0 & 37.4 & 65.5 & 50.3 \\
LoRA & 62.2 & 45.7 & 56.9 & 38.4 & 60.5 & 44.5 & 57.1 & 40.6 & 46.1 & 35.6 & 66.7 & 51.8 \\
DoRA & 61.5 & 45.6 & 56.0 & 37.6 & 58.3 & 41.1 & 55.5 & 41.2 & 45.5 & 36.7 & 66.8 & 51.9 \\
MixLoRA & 61.3 &  44.2 &  61.6 &  34.2 &  61.8 &  41.3 &  55.1 &  40.9 &  45.4 &  36.1 &  66.6 &  53.4 \\
M$^2$PT & 63.7 &  46.2 &  61.0 &  34.0 &  64.3 &  43.2 &  58.0 &  46.8 &  49.7 &  36.3 &  66.1 &  53.1 \\

\hline
\rowcolor{Gray}
\bf \methodname~(Ours) & \bf 75.1 & \bf   54.6 & \bf   67.9 & \bf   38.0 & \bf   67.0 & \bf   45.0 & \bf   79.8 & \bf   56.8 & \bf   54.7 & \bf   42.0 & \bf   75.2 & \bf   60.7 \\
\bottomrule
\end{tabular}
}
\label{tab:translation_main_results}
\end{table}

Table \ref{tab:translation_main_results} shows the image translation performance, measured by the METEOR and BLEU scores, for the six tasks and four datasets.
Our method consistently shows superior performance compared to the baseline methods, highlighting its robustness and generalizability to different image translation scenarios.

\textbf{Comparison to Cascaded Models}: \methodname~significantly outperforms the traditional cascaded OCR + MT pipelines (EasyOCR\_NLLB and PP-OCRv3\_NLLB) on all tasks.
In the task \datasetecoit~(ZH-EN), for example, \methodname~achieves a METEOR score of 75.1 and is thus significantly higher than the scores of 13.7 and 13.1 of the cascaded basic programs.
This illustrates the limitations of relying on independent OCR and MT components, which are prone to error propagation and lack a holistic understanding of the visual and textual context.

\textbf{Comparison to MLLM Baselines}: \methodname~also outperforms the \qwen-3B baseline on all zero-shot, one-shot and chain-of-thought prompting strategies.
Importantly, \methodname~exceeds Text-only on most tasks. This shows that it is able to integrate visual and textual information effectively, even when not relying on a real text.
The improvement over zero-shot, one-shot, and chain-of-thought demonstrates the effectiveness of our method in utilizing training data for targeted performance improvements.

\textbf{Comparison to Fine-tuning Methods}: \methodname~outperforms all compared fine-tuning methods in different image translation datasets, including Full Fine-tuning, LoRA, DoRA, MixLoRA and M$^2$PT.
This underlines the effectiveness of our neuron-aware fine-tuning strategy.
For the \datasetmit~(DE-EN) dataset, for example, \methodname~increases the METEOR score from 62.6 (Full Fine-tuning) to 79.8.
By selectively updating only the most relevant neurons for the image translation task, \methodname~maximizes the benefits of fine-tuning while mitigating the risk of parameter redundancy .

\subsection{Comparison with SOTA IT models}

\begin{table}[t]
    \centering
    \small
    \caption{Quantitative comparison with the SOTA IT models. We conducted a comprehensive evaluation of 6 models  using the test set \datasetecoit~(ZH-EN) and \datasetiitm~(DE-EN) dataset. We highlight the best numbers in \textbf{bold}.}
    \label{tab:translation_sota_results}
    % \resizebox{\textwidth}{!}{
    % \setlength{\tabcolsep}{8pt} % 调整列间距
    \begin{tabular}{lcccc}
        \toprule
        & \multicolumn{2}{c}{\datasetecoit~(ZH-EN)} & \multicolumn{2}{c}{\datasetiitm~(DE-EN)} \\
        \cmidrule(r){2-3} \cmidrule(l){4-5}
        & METEOR & BLEU & METEOR & BLEU \\
        \midrule
        \rowcolor{Gray}
        \bf \methodname~(Ours)  & \bf 75.1 & \bf  54.6 & \bf 67.9 & \bf  38.0 \\
        \midrule
        ItNet & 61.1 & 39.3 & 48.9 & 27.3 \\
        PEIT  & 69.2 & 47.2 & 48.1 & 32.8 \\
        Translatotron-V  & 73.1 & 52.6 & 53.2 & 36.3 \\
        UMTIT & 70.8 & 52.0 & 54.6 & 36.1 \\
        E2ETIT & 46.1 & 31.5 & 32.1 & 21.9 \\
        DIMTDA & 72.4 & 46.6 & 47.9 & 32.4 \\
        \bottomrule
    \end{tabular}
    % }
\end{table}

To further validate the effectiveness of \methodname, we compare its performance with several state-of-the-art image translation (IT) models. 
Table \ref{tab:translation_sota_results} shows the results for the \datasetecoit~(ZH-EN) and \datasetiitm~(DE-EN) datasets.
Our method achieves the best performance for both datasets and outperforms all six compared models in terms of METEOR and BLEU scores. 
In particular, for \datasetecoit~(ZH-EN), \methodname~achieves a METEOR score of 75.1 and a BLEU score of 54.6, significantly outperforming the second best model, Translatotron-V, which scores 73.1 and 52.6 respectively.
In \datasetiitm~(DE-EN), \methodname~achieved a METEOR score of 67.9 and a BLEU score of 38.0 in this German-English translation task, a significant improvement over UMTIT's previous best results of 54.6 (METEOR) and 36.1 (BLEU).

These results show that our method is able to outperform existing specialized models for different language pairs. 
The superior performance can be attributed to the neuron-aware fine-tuning that effectively utilizes the multimodal capabilities of large language models while preserving the pre-trained knowledge and adapting to the specific nuances of the image translation task.

\subsection{Ablation study}

%%%%%%%%%%%%%%%%%%%%%%%%%%%%%%%%%%%%%%%%%%%%%%%%%%%%%%%
\begin{table}[t]
    \centering
    \small
    \caption{Results of the ablation study comparing our method with different model variants on \datasetopus~(EN-ZH) and \datasetecoit~(ZH-EN).}
    \renewcommand{\arraystretch}{1.1}
% \resizebox{\columnwidth}{!}{
\begin{tabular}{lcc|cc}
    \toprule
    & \multicolumn{2}{c|}{\datasetopus~(EN-ZH)} & \multicolumn{2}{c}{\datasetecoit~(ZH-EN)} \\
    \cmidrule(lr){2-3} \cmidrule(lr){4-5}
    & METEOR & BLEU & METEOR & BLEU \\
    \midrule
    \rowcolor{Gray}
    \bf \methodname~(Ours) & \bf   75.2 & \bf   60.7 & \bf  \bf 75.1 & \bf   54.6 \\
    \hline
    w / General Neurons FT & 65.9 &  51.7 & 51.8 &  29.1 \\
    w / Specific Neurons FT & 66.5 &  52.8 & 61.7 &  42.7 \\
    \hline
    w / ALL Layers FT & 67.2 &  53.3 & 61.6 &  45.8 \\
    w / Language Layers FT& 66.5 &  51.0 & 58.7 &  42.3 \\
    w / Vision Layers  FT& 65.9 &  50.8 & 57.3 &  38.2 \\
    \bottomrule
\end{tabular}
% }

\label{tab:ablation}
\end{table}
%%%%%%%%%%%%%%%%%%%%%%%%%%%%%%%%%%%%%%%%%%%%%%%%%%%%%%%

To investigate the contribution of the different components of \methodname, we conducted an ablation study using the \datasetopus~(EN-ZH) and \datasetecoit~(ZH-EN) datasets. 
The results are shown in Table \ref{tab:ablation}. 
We analyzed the effects of fine-tuning different neuron groups and layers:

\noindent \textbf{Effect of neuron type}:
\begin{itemize}
    \item General Neurons FT: Fine-tuning only general neurons in both language and vision layers resulted in a significant drop in performance compared to \methodname. 
    This suggests that general neurons alone are not sufficient to achieve optimal performance in image translation. 
    They likely capture broader features but lack the specific knowledge required for accurate translation.
    \item Specific Neurons FT: Fine-tuning only specific neurons produced significantly better results than fine-tuning general neurons and approached the performance of \methodname, especially on the \datasetecoit~dataset. 
    This highlights the crucial role of specific neurons in capturing the nuanced information necessary for image translation. 
    These neurons are likely specialized for linguistic features of the target language or the visual representation of text.
\end{itemize}

\noindent \textbf{Effect of Layer Selection}:
\begin{itemize}
    \item All Layers FT (without neuron selection): Fine-tuning all neurons in both the language and visual layers, which is essentially \methodname~without the neuron selection mechanism, resulted in lower performance compared to the full \methodname~method. 
    This demonstrates the importance of the neuron selection strategy to improve the effectiveness of the fine-tuning and prevent potential negative effects of updating irrelevant parameters.
    \item Language Layers FT: Fine-tuning only the language layers, which includes both general and specific neurons, performed comparably well to the FT method for specific neurons in the \datasetopus~dataset, but worse in \datasetecoit. 
    This suggests that the contribution of visual layer adaptation is more pronounced in some translation tasks.
    \item Vision Layers FT: Fine-tuning only the visual layers resulted in the lowest performance among the ablation settings. 
    This suggests that while adjusting the visual features is beneficial, the primary performance gain likely comes from adjusting the language layers that are directly responsible for producing the translated text.
\end{itemize}

This ablation study confirms the effectiveness of the core components of our method. 
The selection of specific neurons plays a crucial role in achieving optimal performance. 
While fine-tuning all layers provides some improvement over baseline MLLMs, the targeted approach of \methodname~shows superior performance. 
The results also suggest that the relative importance of matching vision and language layers may vary depending on the specific language pair and the characteristics of the dataset. 
The combination of specific neuron selection and joint vision-language fine-tuning leads to the best overall performance.

\section{Analysis}
%%1 分析泛化能力
% \subsection{Transfer Learning}
%%2 每层重要神经元分数、以及delta分数
\subsection{Neuron identification analysis}

% %%%%%%%%%%%%%%%%%%%%%%%%%%%%
\begin{figure}[t]
    \centering
    
    \subfloat[]{
        \includegraphics[width=0.4\textwidth]{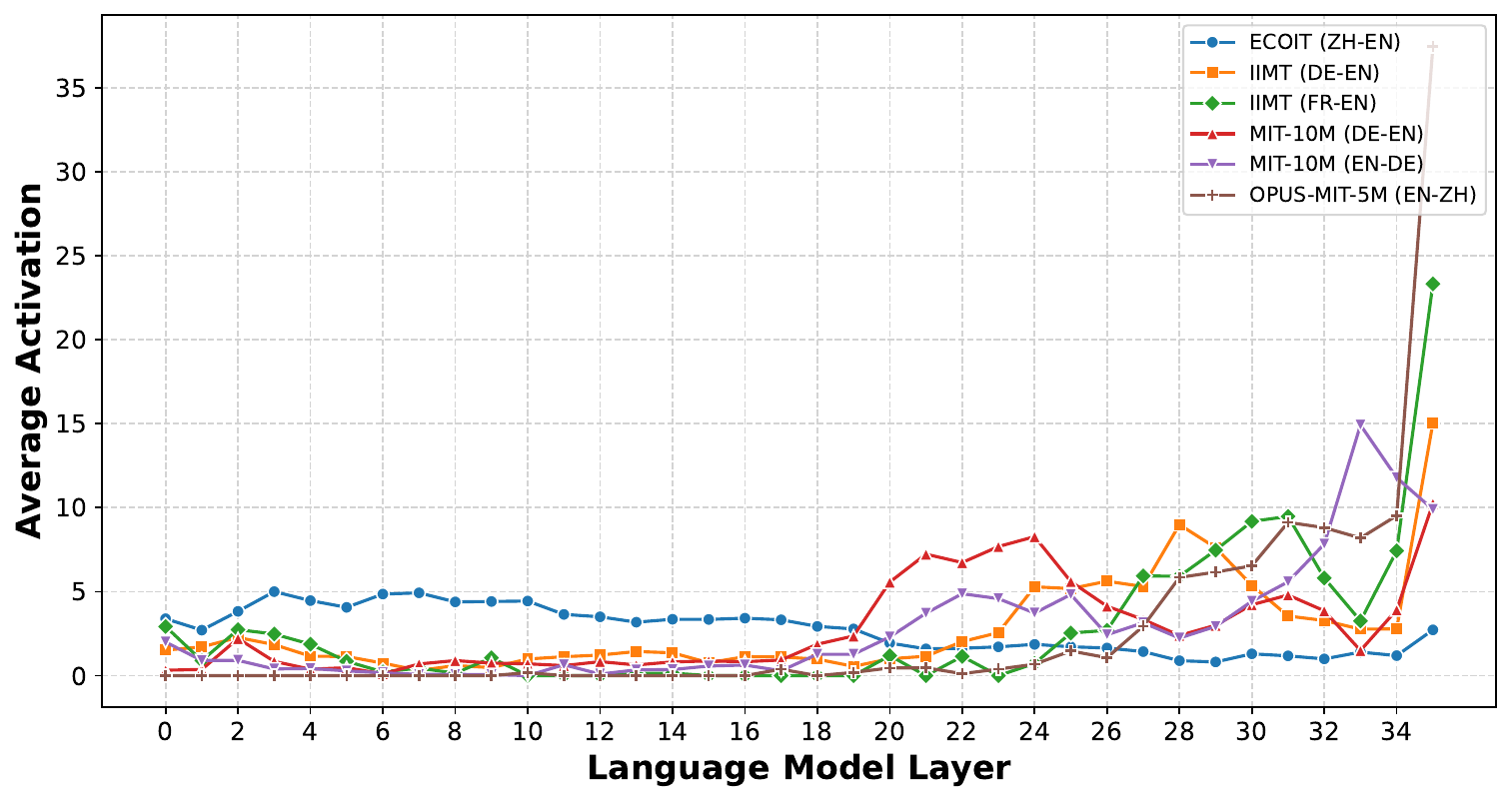}}
    \subfloat[]{
        \includegraphics[width=0.4\textwidth]{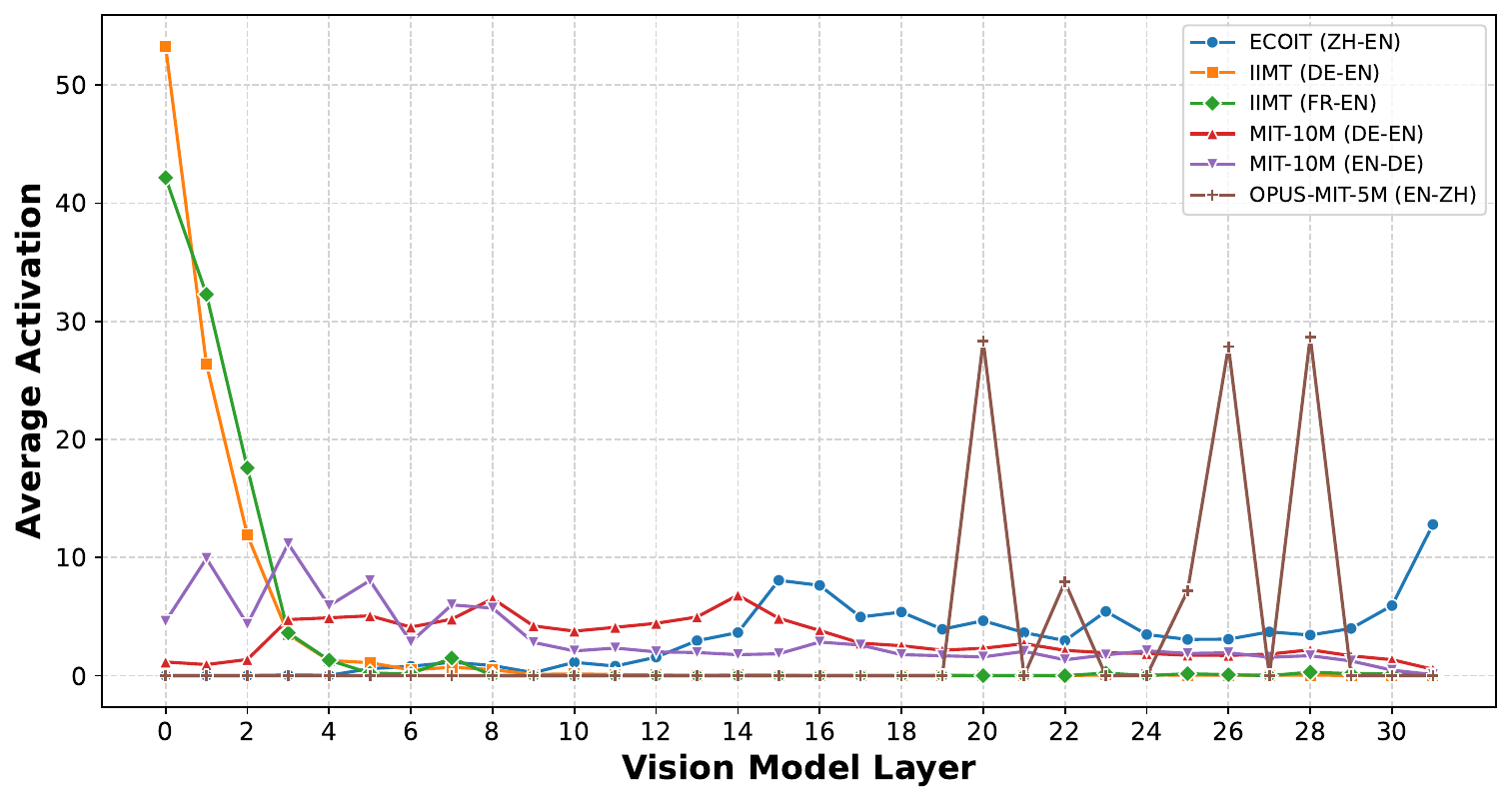}}
    \caption{Average Activation. (a) Language Model Layer Average Activation. (b) Vision Model Layer Average Activation. }
\label{fig:aa}
\end{figure}
%%%%%%%%%%%%%%%%%%%%%%%%%%%%

% %%%%%%%%%%%%%%%%%%%%%%%%%%%%
\begin{figure}[t]
    \centering
    
    \subfloat[]{
        \includegraphics[width=0.4\textwidth]{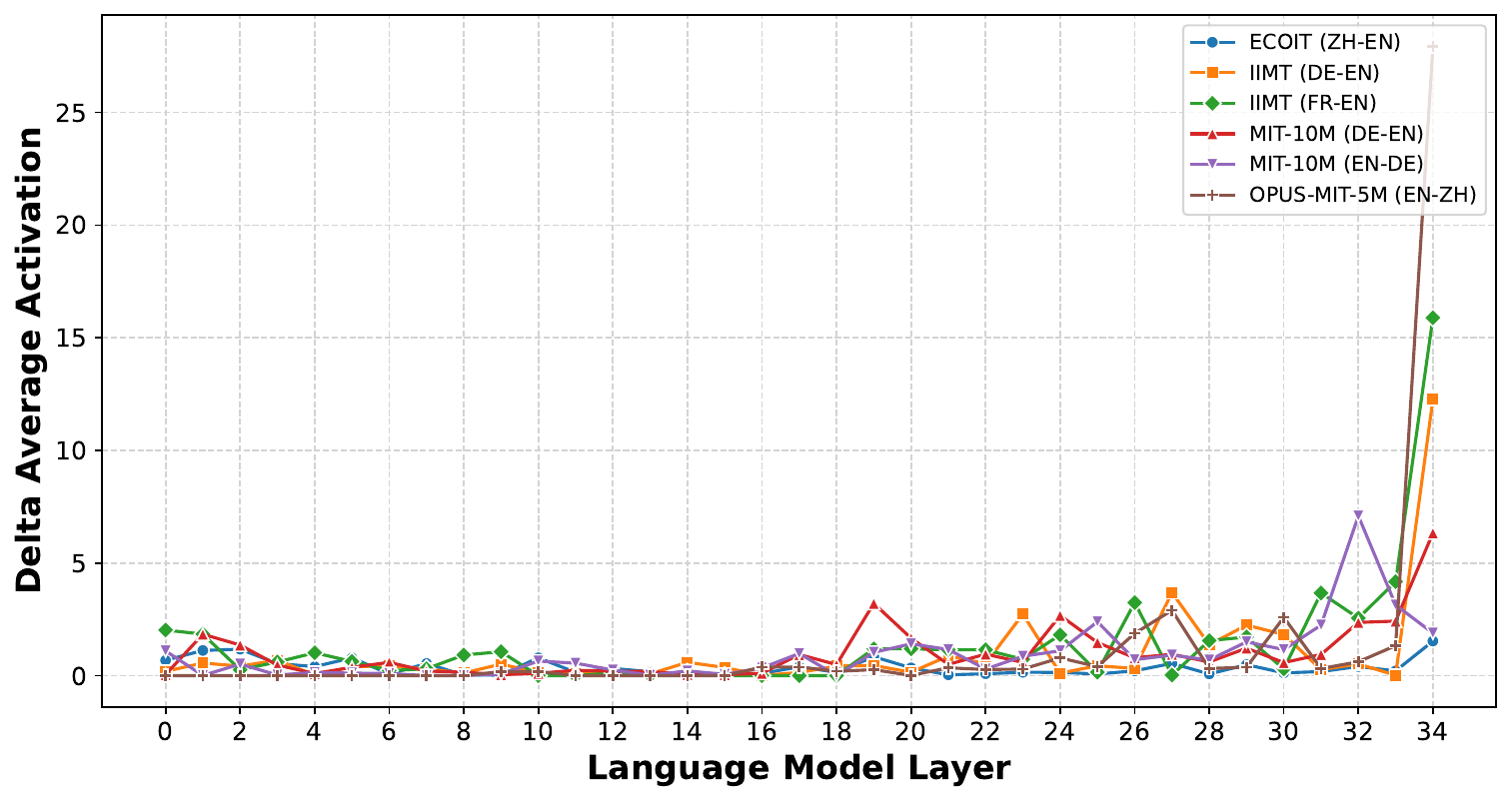}}
     \subfloat[]{
        \includegraphics[width=0.4\textwidth]{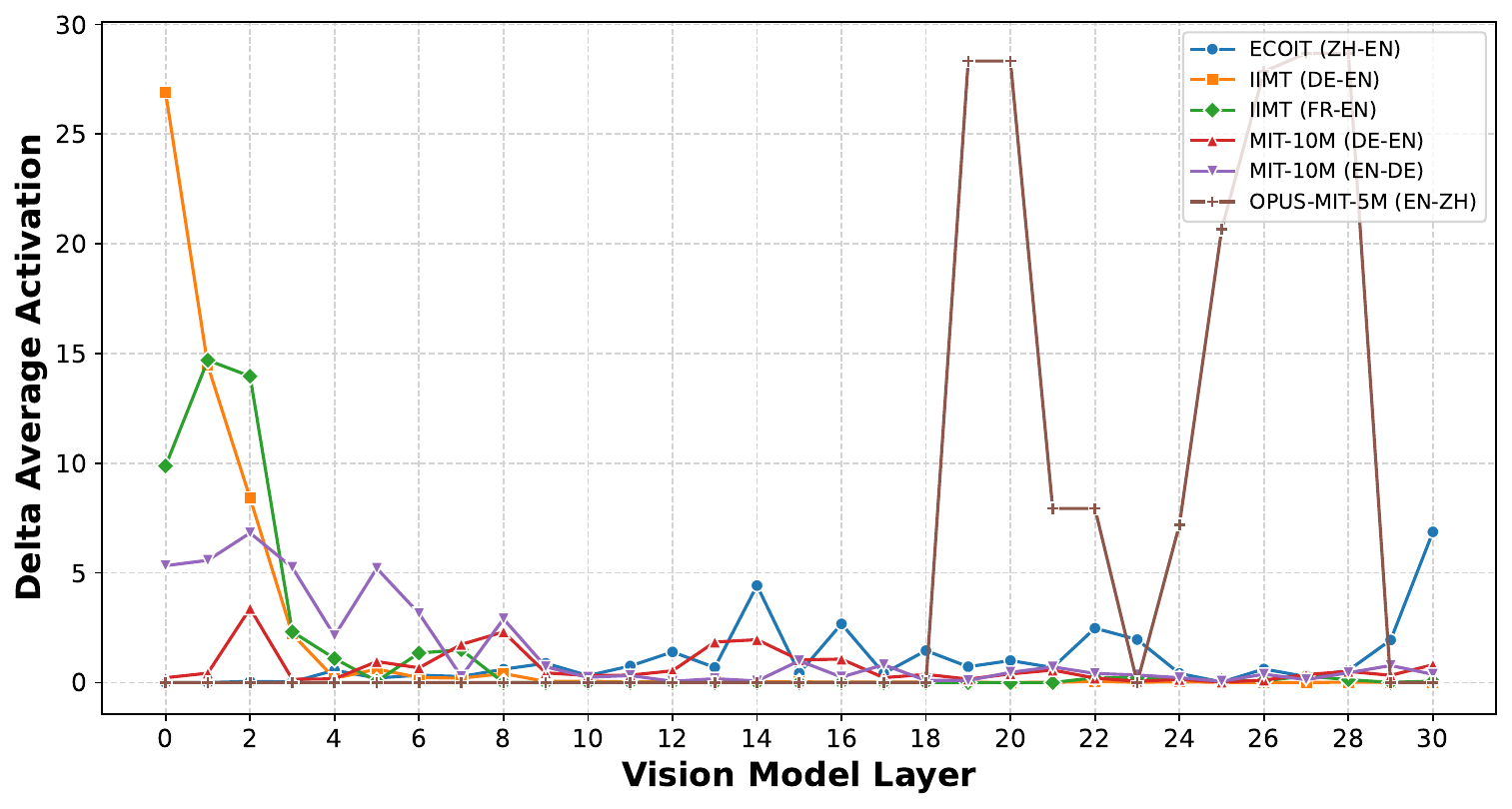}}
    \caption{Delta Average Activation. (a) Language Model Layer Delta Average Activation. (b) Vision Model Layer Delta Average Activation. }
\label{fig:daa}
\end{figure}
%%%%%%%%%%%%%%%%%%%%%%%%%%%%

Figures~\ref{fig:aa} and \ref{fig:daa} illustrate the average neuron activation and the average delta activation for each task for both the language model and vision model layers.
Figure~\ref{fig:aa} shows which layers and neuron groups show higher overall activation in different tasks. 
For example, the \datasetopus~(EN-ZH) task shows significantly higher activation in specific layers of the vision model, suggesting that these layers are particularly sensitive to the visual features present in this dataset. 
Similarly, certain layers of the language model show higher activation for the \datasetecoit~(ZH-EN) task. 
This difference in activation patterns between tasks is evidence of specialization within the model.
Figure~\ref{fig:daa} shows the change in activation between successive layers. 
High delta values indicate layers where activation changes drastically, possibly indicating important processing steps or transitions between different levels of representation. 
The peaks observed for different tasks in different layers indicate task-specific processing within the model. 
For example, the sharp peak for the \datasetopus~dataset in the vision model suggests a significant shift in representation within that specific layer, possibly related to the processing of visual features unique to that task.

These visualizations in combination with our neuron selection method provide insight into how different parts of the model contribute to different image translation tasks. 
The identification of language-specific neurons and the observation of task-dependent activation patterns support the hypothesis that different neurons and layers are specialized to handle different aspects of the image translation process. 
This understanding motivates our neuron-based fine-aware fine-tuning strategy, which enables a more effective and targeted adaptation of the model to specific tasks.

%%3 不用任务下 专有和通用的分数
\subsection{Clustering analysis of general and specific neurons}
%%4 tnse可视化图 
% \subsection{Language Cluster}

% %%%%%%%%%%%%%%%%%%%%%%%%%%%%
\begin{figure}[t]
    \centering
    \subfloat[]{
        \includegraphics[width=0.4\textwidth]{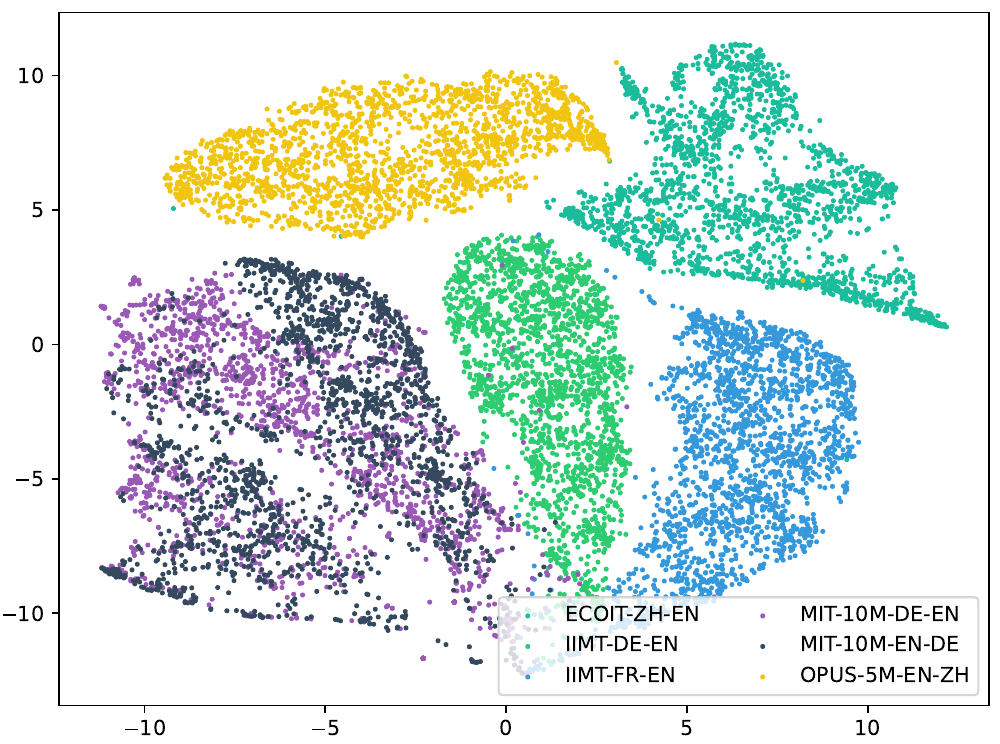}}
    \subfloat[]{
        \includegraphics[width=0.4\textwidth]{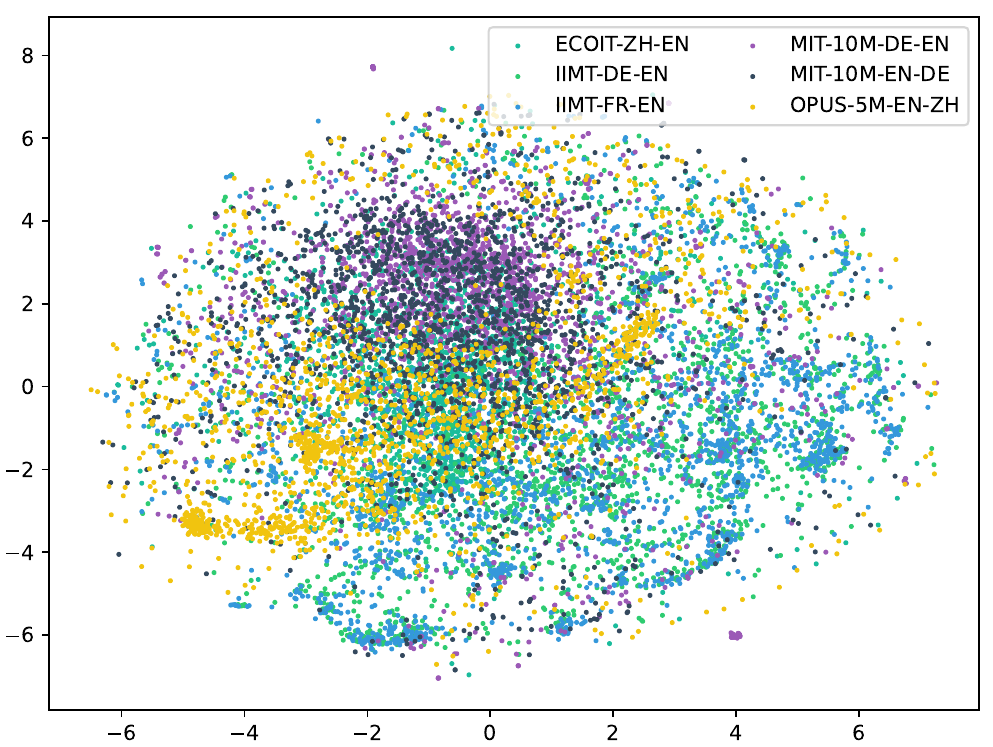}}
        
    \subfloat[]{
        \includegraphics[width=0.4\textwidth]{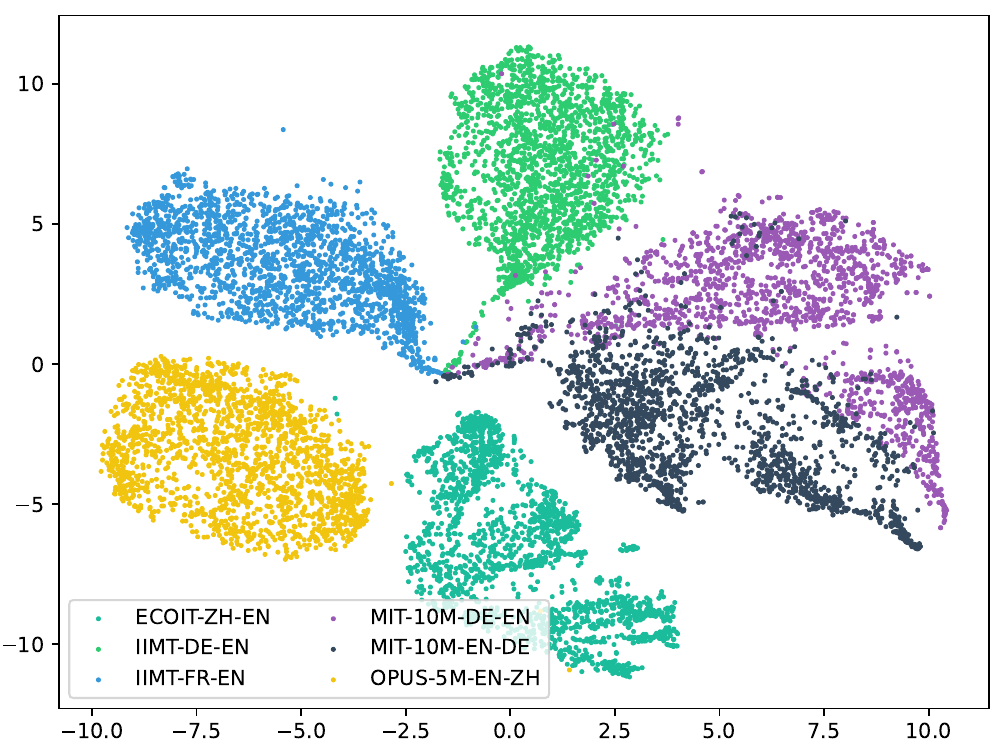}}
    \subfloat[]{
        \includegraphics[width=0.4\textwidth]{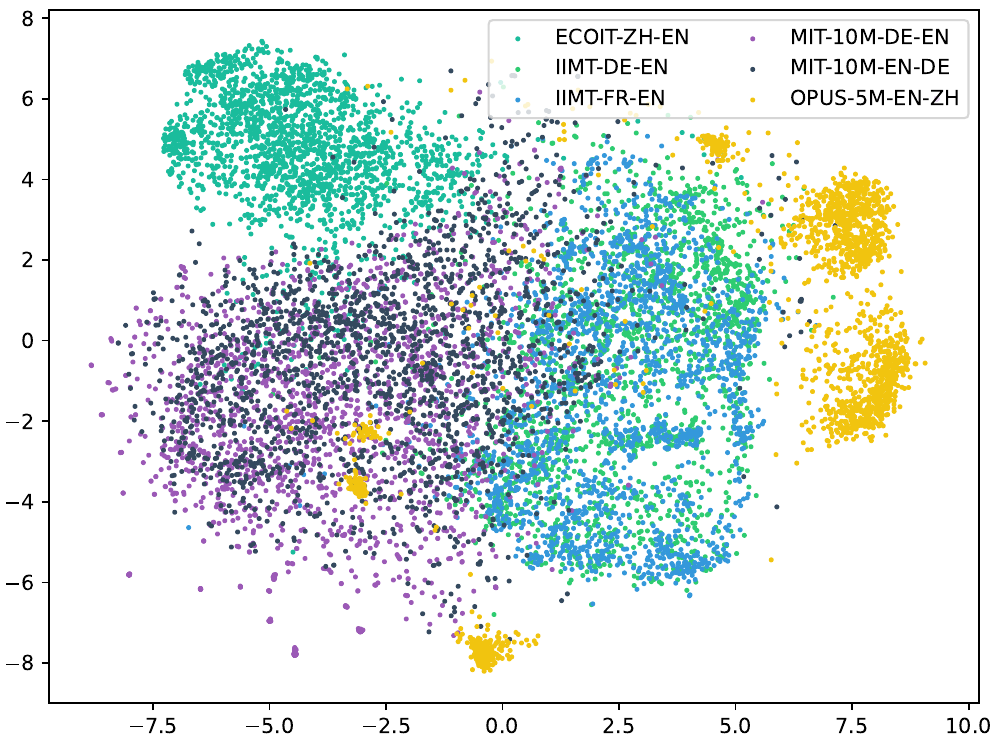}}
\caption{Clustering of representations. (a) Language Model Layer specific neurons. (b) Language Model Layer general neurons. (c) Vision Model Layer specific neurons. (d) Vision Model Layer general neurons.}
\label{fig:Clustering_of_representations}
\end{figure}
%%%%%%%%%%%%%%%%%%%%%%%%%%%%

To further investigate the role of general and specific neurons in capturing language and visual knowledge for image translation, we used t-SNE to visualize the representations learned by these neurons across different tasks. Figure~\ref{fig:Clustering_of_representations} shows the t-SNE plots of neuron activations from the last layer of both vision and language modules.

Analysis of Language Model Neurons (Figures \ref{fig:Clustering_of_representations} (a) and (b) ):
\begin{itemize}
    \item \textbf{Specific Neurons (a)}: The t-SNE diagram of language-specific neurons shows clear clusters corresponding to different tasks of image translation. 
    This clear separation indicates that these neurons are specialized in capturing task-specific linguistic features. The clusters for tasks with similar languages (e.g. DE-EN and FR-EN) appear to be closer to each other than those with different languages (e.g. ZH-EN and EN-ZH), supporting our hypothesis that language-specific neurons learn linguistic nuances relevant to specific languages and translation directions.
    \item  \textbf{General Neurons (b)}: In contrast, general neurons show a more mixed distribution. 
    Although some loose groupings are evident, the lack of clear separation suggests that they capture more general task-related linguistic features, consistent with our expectation that they encode broader linguistic knowledge that applies to a wider range of languages.
\end{itemize}

Analysis of Vision Model Neurons (Figures \ref{fig:Clustering_of_representations} (c) and (d) ):
\begin{itemize}
    \item \textbf{Specific Neurons (c)}: Similar to the language model, the specific neurons of the vision model also show distinct, albeit less defined, task-related clusters, indicating specialization in task-relevant visual features. 
    The less distinct clustering suggests that visual features are less strongly associated with individual languages than with linguistic features.
    \item \textbf{General Neurons (d)} The general neurons in the vision model show a more diffuse distribution, suggesting that they capture general, task-related visual features, such as recognition of text regions or image layout. 
    The less defined clustering supports the idea that general visual features in image translation are less task-specific than language-related features.
\end{itemize}

The t-SNE visualizations provide strong evidence for the specialization of specific neurons in the detection of task-related linguistic and visual features. 
This is consistent with the core motivation of \methodname, which selectively tunes these neurons for improved task performance. 
The diffuse distribution of general neurons suggests that they play a role in capturing broader, task-related knowledge that is retained by our selective fine-tuning. 
These results validate the design choices of \methodname~and provide insights into the distinct roles of different neuron types in large multimodal models of image translation.

%%5 通用性：qwen对比
\subsection{Results on \qwen-7B and \llavanext}

% %%%%%%%%%%%%%%%%%%%%%%%%%%%%
\begin{figure}[t]
    \centering
    \subfloat[]{
        \includegraphics[width=0.4\textwidth]{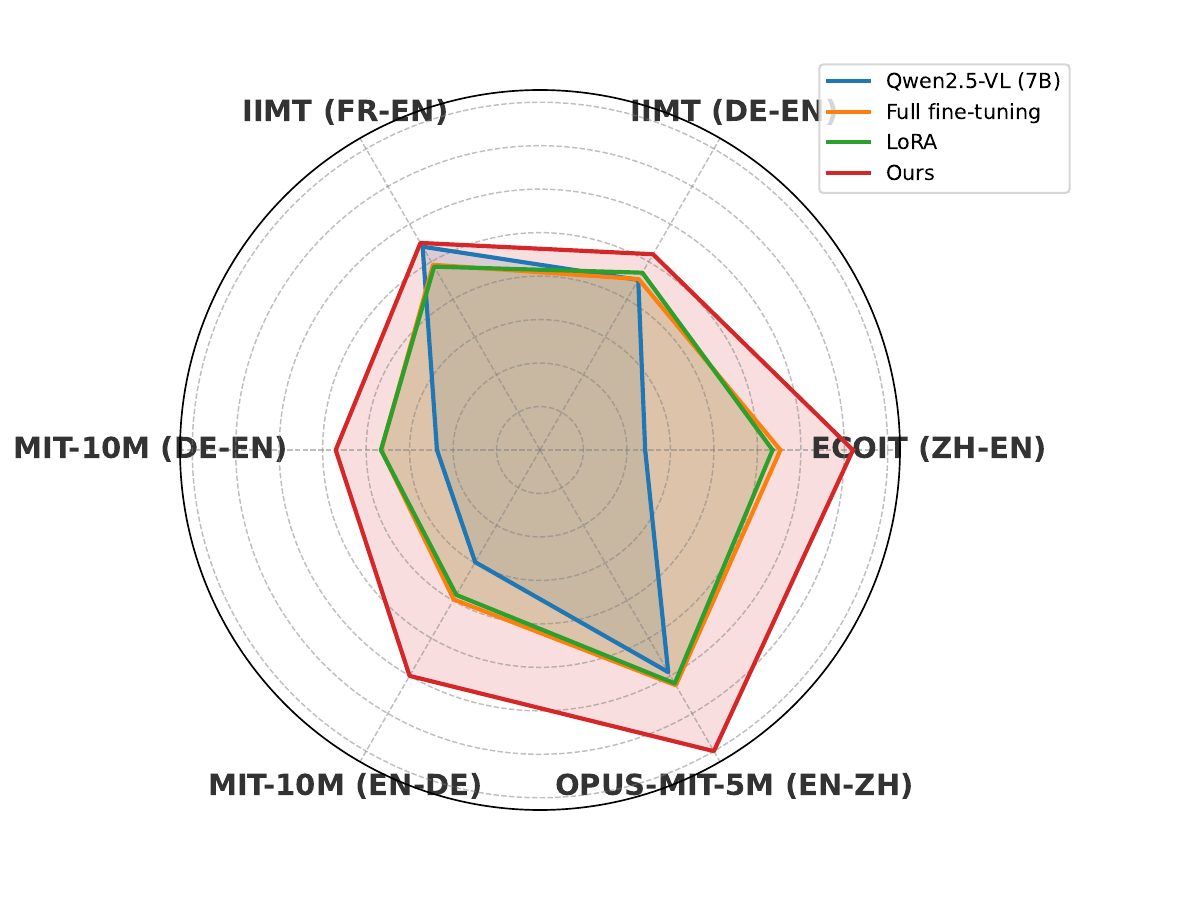}}
    \subfloat[]{
        \includegraphics[width=0.4\textwidth]{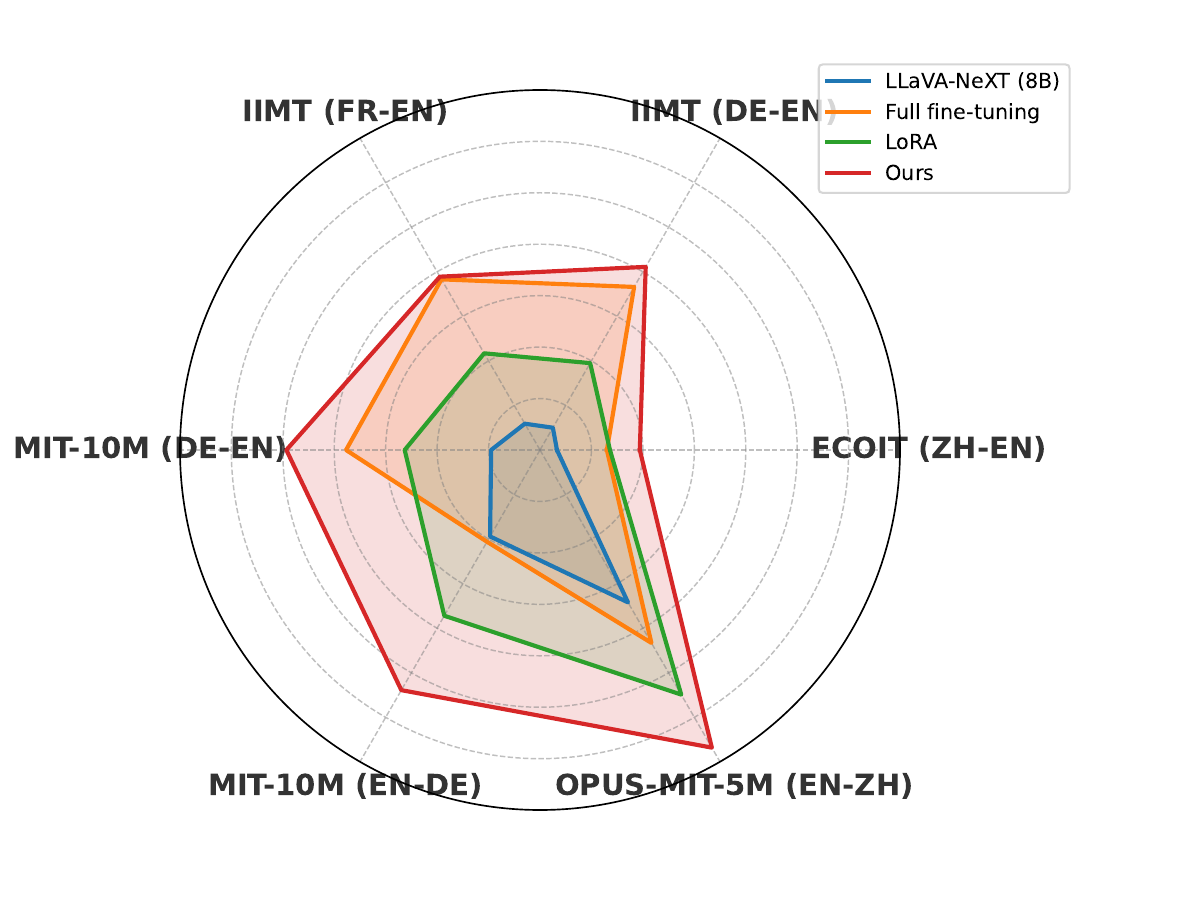}}
\caption{
Comparison of METEOR with six datasets for fine-tuning the \qwen-7B (a) and \llavanext-\llama~(8B) (b) using Full Fine-tuning, LoRA and \methodname~(Ours).}
\label{fig:other_model_generalization}
\end{figure}
%%%%%%%%%%%%%%%%%%%%%%%%%%%%

To evaluate the generalizability and scalability of our method, we conducted experiments with larger models: \qwen-7B and 
\llavanext-\llama~(8B). 
Figure~\ref{fig:other_model_generalization} shows the consistent effectiveness of our method with different model architectures and sizes. 
The figure shows the METEOR scores obtained on the six image translation tasks for each model and fine-tuning method.

For the \qwen-7B model, \methodname~shows significant performance improvements compared to Full Fine-tuning and LoRA on all tasks. 
The figure shows that the \methodname~range significantly outperforms that of the other methods, indicating its overall superior performance. 
Although both Full Fine-tuning and LoRA improve the baseline performance of \qwen-7B (represented by the innermost hexagon), they fall well short of our method. 
This underscores the effectiveness of our neuronal fine-aware fine-tuning strategy in utilizing the increased capacity of the larger model.
The experiments with \llavanext~further confirm the generalizability of our method. 
Although the overall METEOR score is lower for this model compared to \qwen-7B, \methodname~outperforms both Full Fine-tuning and LoRA on all tasks. 
The figure for \llavanext~shows a similar trend to the \qwen-7B experiments, with \methodname~covering a wider range than the other methods.

The consistent improvements observed in both \qwen-7B and \llavanext~demonstrate the robustness and scalability of \methodname. 
Our neural fine-tuning strategy appears to be particularly effective in exploiting the increased capacity of larger models, leading to more significant gains compared to standard fine-tuning and LoRA. 
The figure clearly illustrates this superior performance on various translation tasks. 
These results suggest that our method offers a promising approach to maximizing the performance of MLLMs in IT tasks, regardless of the specific model architecture or scale.

\subsection{Case study}
%%%%%%%%%%%%%%%%%%%%%%%%%%%%%%%%%%%%%%%%%%
\begin{figure}[t]
  \centering
   \includegraphics[width=0.8\textwidth]{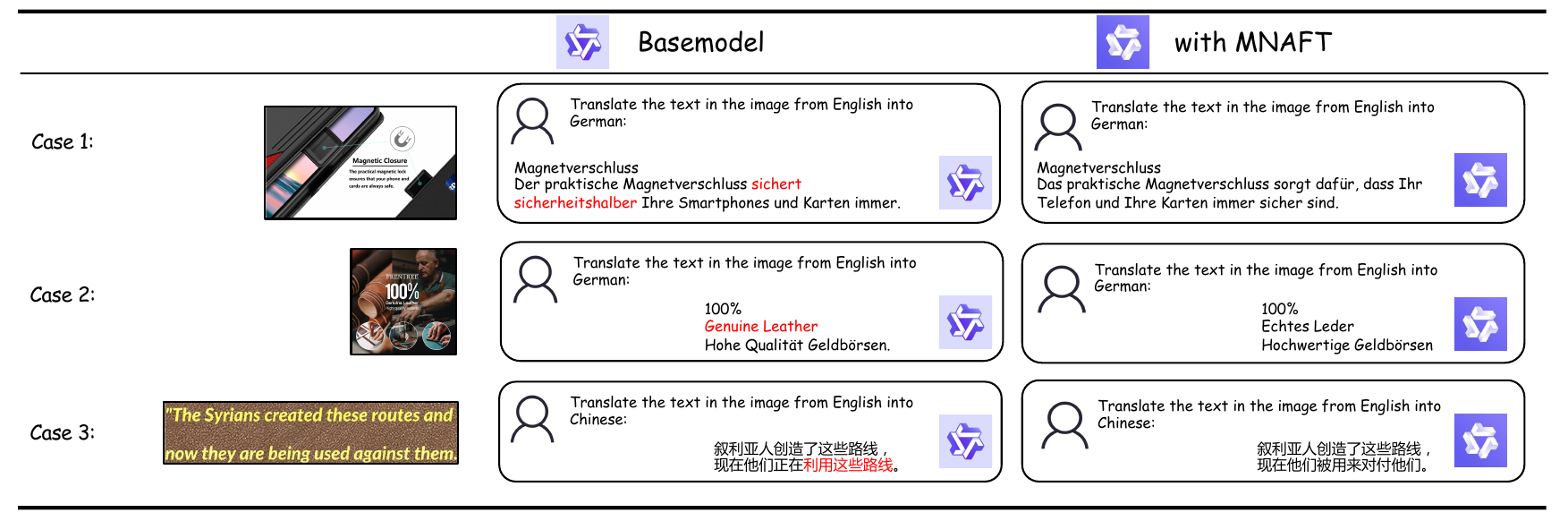}
  \caption{Case Study. }
  \label{fig: casestudy}
\end{figure}
%%%%%%%%%%%%%%%%%%%%%%%%%%%%%%%%%%%%%%%%%%

To further illustrate the practical benefits of \methodname, we present a qualitative analysis of its performance using specific examples. 
As shown in Figure \ref{fig: casestudy}, these case studies show how \methodname~leads to more accurate and contextually appropriate translations.

\textbf{Case 1:} This case shows an image with product packaging that contains both English text and a stylized logo. 
The base model (\qwen-3B) hallucinates additional text, ``sicherheitshalber Ihre Smartphones und Karten immer'', translating to ``always your smartphones and cards for safety''), which may be influenced by common product descriptions for cell phone cases. 
In contrast, \methodname~correctly translates the entire sentence as ``Das praktische Magnetverschluss sorgt dafür, dass Ihr Telefon und Ihre Karten immer sicher sind''. 
This demonstrates the robustness of \methodname~and its ability to capture the full meaning in the visual context.

\textbf{Case 2:}  This case shows a promotional image with the text ``100\% Genuine Leather. 
High quality wallets.'' The base model provides a literal translation of ``Hohe Qualität Geldbörsen'' (``High quality wallets'' for the second sentence. While this is grammatically correct, it misses the nuance of ``High Quality Wallets'' which is often understood as a product category. 
\methodname~correctly identifies and translates ``Genuine Leather'' as ``Echtes Leder'' and provides a more natural translation of ``Hochwertige Geldbörsen'' (``High-quality wallets''), demonstrating a deeper understanding of the implied meaning within the advertising context.

\textbf{Case 3:}  This case is a headline image with the sentence ``The Syrians created these routes and now they are being used against them.'' 
The base model translates this into Chinese which means that ``The Syrians created these routes and now they are using these routes.'' This translation loses the crucial context of the sentence, in which ``they'' in the second part of the sentence refer to the routes being used against the Syrians. 
\methodname~accurately captures this nuance and translates it, which correctly reflects that the routes are used ``against the Syrians''. 
This demonstrates \methodname's superior contextual understanding and disambiguation capabilities.

These case studies qualitatively demonstrate the advantages of \methodname~in tackling challenging scenarios in image translation. 
By selectively fine-tuning relevant neurons, \methodname~achieves a better balance between translation accuracy, visual fidelity, and contextual understanding, making it a promising approach for real-world image translation applications.

%%%%%%%%%%%%%%%%%%%%%%%%%%%%%%%%%%%%%%%%%%%%%%%%%%%%%%%
\begin{table}[t]
    \centering
    \caption{Computational cost comparison on the \datasetecoit~and \datasetopus~datasets. Time is reported in hours (h) and GPU memory in gigabytes (GB). Experiments are conducted on NVIDIA A100 (80GB) GPUs. 
    The neuron identification stage of \methodname~introduces only a negligible one-time cost 
    (about 2 minutes for \datasetecoit~and 3 minutes for \datasetopus), which is not included in the table. All methods are trained for the same number of epochs and batch size.}
    \label{tab:computational_cost}
    % \resizebox{\columnwidth}{!}{%
    \renewcommand{\arraystretch}{1.1}
    \begin{tabular}{l|cc|cc}
        \toprule
        \multirow{2}{*}{\textbf{Method}} & \multicolumn{2}{c|}{\textbf{ECOIT}} & \multicolumn{2}{c}{\textbf{OPUS-MIT-5M}} \\
        \cmidrule(lr){2-3} \cmidrule(lr){4-5}
        & \textbf{Time (h)} & \textbf{Memory (GB)} & \textbf{Time (h)} & \textbf{Memory (GB)} \\
        \midrule
        Full Fine-tuning & 9.2 & 94.3 & 12.6 & 127.7 \\
        LoRA & 8.6 & 12.2 & 10.2 & 12.3 \\
        \rowcolor{Gray} \methodname~(Ours) & \textbf{7.0} & 20.8 & \textbf{8.2} & 20.8 \\
        \bottomrule
    \end{tabular}%
    % }
\end{table}
%%%%%%%%%%%%%%%%%%%%%%%%%%%%%%%%%%%%%%%%%%%%%%%%%%%%%%%

\subsection{Computational cost analysis}
\label{sec:cost_analysis}
To evaluate the training efficiency of our proposed \methodname, 
we conducted a detailed comparison with Full Fine-tuning and LoRA in terms of wall-clock training time and peak GPU memory usage. 
The results on the \datasetecoit~and \datasetopus~datasets are summarized in Table~\ref{tab:computational_cost}.
The training pipeline of \methodname~comprises two stages: neuron identification and selective fine-tuning. 
The identification stage requires computing neuron importance scores once on a small scoring set. 
On \datasetecoit~and \datasetopus, this step is completed within $2$ minutes and $3$ minutes, respectively, 
which is negligible compared to the overall training time (on the order of hours). 
Therefore, the efficiency of \methodname~is dominated by the subsequent fine-tuning phase.

Compared with Full Fine-tuning, \methodname~demonstrates substantial efficiency gains. 
In \datasetecoit, its fine-tuning is $24\%$ faster (7.0 h vs. 9.2 h),
while consuming only $22\%$ of the GPU memory (20.8 GB vs. 94.3 GB). 
Similar improvements are observed in \datasetopus. 
These results empirically validate our design of selectively updating a small subset of neurons to achieve significant reductions in training cost.
The comparison with LoRA reveals an interesting trade-off. 
LoRA is the most memory-efficient method (about 12GB), as it only introduces a small number of low-rank adaptation matrices. 
\methodname~uses moderately more memory (about 21GB), 
because it must maintain optimizer states for a subset of the original parameters. 
Nevertheless, \methodname~achieves noticeably faster training:
its fine-tuning phase is $18\%$ faster on \datasetecoit~and $20\%$ faster on \datasetopus. 
We attribute this advantage to LoRA’s additional forward-pass operations (extra matrix multiplications), 
which introduce latency, while \methodname~retains the original architecture. 
Our method applies gradient masking during backpropagation, 
so optimizer updates are restricted to the selected neurons. 
Although gradients are still propagated for all activations in the current implementation, 
skipping parameter updates for frozen neurons reduces effective computation and leads to shorter training time.

Overall, \methodname~offers a favorable balance between time and memory:
it is substantially more efficient than Full Fine-tuning and faster than LoRA while maintaining manageable memory overhead. 
However, we note that the current implementation does not prune the gradient computation itself—masking is applied at the optimizer update stage. 
As a result, further efficiency gains may be possible with specialized implementations that support sparse backpropagation. 
We leave such optimizations for future work.

%%%%%%%%%%%%%%%%%%%%%%%%%%%%%%%%%%%%%%%%%%%%%%%%%%%%
\section{Broader applicability and future work}

While the efficacy and efficiency of MNAFT have been thoroughly established within the domain of image translation, the underlying architectural insights and adaptive mechanisms are designed with broader applicability in mind. 
This section delves into the inherent versatility of MNAFT, exploring its potential to enhance performance across a wider spectrum of multimodal tasks and outlining future investigative pathways.

\subsection{The foundation for generalized adaptability}

MNAFT's operational premise rests on the observation that within expansive MLLMs, distinct neuronal populations assume specialized functions during their extensive pre-training. 
These specializations manifest in processing specific sensory inputs, linguistic structures, or intricate inter-modal relationships. 
This functional segregation is a universal characteristic of how complex neural architectures learn from diverse data, rather than being exclusive to image-to-text conversion. 
Our methodology, which integrates instruction-guided activation profiling with Taylor expansion-derived salience scores, provides a potent framework for:
\begin{itemize}
    \item  Identifying specific aggregations of neurons that demonstrate heightened activity or sensitivity when confronted with particular data types (e.g., visual attributes, grammatical constructs, emotional undertones) or in the context of specific assignments.
    
    \item Evaluating the precise impact these specialized neuronal units have on a given downstream objective, thereby enabling highly targeted parameter adjustments.

    \item By strategically preserving the weights of non-essential neurons and layers, MNAFT effectively counteracts common challenges such as catastrophic forgetting and the redundant updating of parameters—issues frequently encountered when adapting large, foundation models.
\end{itemize}

\subsection{Envisioning applications beyond textual image conversion}
We foresee substantial advantages in applying MNAFT's neuron-aware fine-tuning paradigm to a variety of other multimodal challenges:

\begin{itemize}
    \item Crafting articulate descriptions for images demands both acute visual perception and fluent linguistic expression. 
    MNAFT could focus on neurons responsible for detecting salient objects, dynamic actions, and their contextual relationships, alongside those crucial for constructing grammatically sound and contextually appropriate narratives. 
    This would be particularly valuable for specialized domains (e.g., medical diagnostics, fashion commentary) or for generating multilingual captions.

    \item For visual question answering (VQA), it necessitates a comprehensive understanding of visual scenes coupled with the interpretation of natural language queries to formulate accurate text-based responses. 
    MNAFT could be instrumental in isolating neurons dedicated to object recognition, property extraction, spatial reasoning, and query comprehension.
    For instance, specific neuron clusters might be optimized for discerning ``color" attributes, while others excel at handling ``quantity" inquiries.
    Targeted fine-tuning could sharpen the model's ability to respond to particular categories of visual questions with greater precision, bypassing the need for wholesale model retraining.

    \item In human-AI interaction, MLLMs must process visual information concurrently with natural language conversation. 
    MNAFT could help customize the model for different dialog scenarios, such as locating objects referenced in discourse, inferring user intent from visual cues, or formulating visually grounded replies.
    Neuronal subsets specialized in dialog coherence or sentiment recognition could be precisely updated.
\end{itemize}

%%%%%%%%%%%%%%%%%%%%%%%%%%%%%%%%%%%%%%%%%%%%%%%%%%%%%%
\section{Conclusion}

In this paper, we present \methodname, a novel neuron-aware fine-tuning method to improve the performance of multimodal large language models (MLLMs) for image translation. 
\methodname~exploits the insight that different neurons within MLLMs play different roles in the processing of multimodal information, some specialized for specific languages, and others capture general or cross-modal knowledge. 
By selectively fine-tuning only the most relevant neurons for a given image translation task, \methodname~maximizes the benefits of adaptation while mitigating the risk of parameter redundancy.
Our extensive experiments with different datasets and language pairs have shown that \methodname~consistently outperforms existing SOTA IT methods, including traditional cascade pipelines and various LLM fine-tuning strategies. 
The ablation studies confirmed the effectiveness of our neuron selection mechanism and the importance of joint vision-language adaptation. 
Further analyses using visualizations provided convincing evidence for the specialization of different neuron types within the model, supporting the underlying principles of \methodname.
The successful application of \methodname~to larger models such as \qwen-7B and \llavanext~demonstrated its scalability and generalizability in different MLLMs architectures.
In the next step, we will investigate the transferability of the learned neuron specializations to different tasks and modalities. Applying \methodname~to other complex multimodal tasks, such as visual question answering or image captioning, could unlock significant performance improvements in these domains.

%%%%%%%%%%%%%%%%%%%%%%%%%%%%%%%%%%%%%%%%%%%%%%%%%%%%%%%
%%% Acknowledgements. 致谢
%%%%%%%%%%%%%%%%%%%%%%%%%%%%%%%%%%%%%%%%%%%%%%%%%%%%%%%
% \Acknowledgements{This work was supported by the National Natural Science Foundation of China (Grant Nos. 00000000 and 11111111).}
\Acknowledgements{We thank the anonymous reviewers for their insightful comments and suggestions.
This work was supported by the National Key Research and Development Program of China (Grant Nos. 2024YFB3309702, 2023YFE0116400) and the National Natural Science Foundation of China Youth Fund (Grant No. 62306210).}

% \bibliographystyle{scis}
% \bibliography{refs}

\end{document}